\documentclass[10pt,twocolumn,letterpaper]{article}

\usepackage{cvpr}              
\usepackage{multirow}
\usepackage{rotating}
\usepackage{tabto}

\usepackage[dvipsnames]{xcolor}

\definecolor{cvprblue}{rgb}{0.21,0.49,0.74}
\usepackage[pagebackref,breaklinks,colorlinks,citecolor=cvprblue]{hyperref}
\usepackage[accsupp]{axessibility}

\def\paperID{9805} 
\def\confName{CVPR}
\def\confYear{2024}

\newcommand{\term}{self-/unsupervised }
\newcommand{\Term}{Self-/unsupervised }

\title{Mining Supervision for Dynamic Regions in Self-Supervised Monocular \\ Depth Estimation}

\author{
\text{Hoang Chuong Nguyen}\textsuperscript{1} \hspace{0.5cm} \text{Tianyu Wang}\textsuperscript{1} \hspace{0.5cm}  \text{Jose M. Alvarez}\textsuperscript{2}  \hspace{0.5cm}
\text{Miaomiao Liu}\textsuperscript{1}\\
\textsuperscript{1}\text{Australian National University} \hspace{1.0cm}  \textsuperscript{2}\text{NVIDIA}\\
{\tt\small {\{hoangchuong.nguyen, tianyu.wang2, miaomiao.liu\}@anu.edu.au } \hspace{0.5cm} josea@nvidia.com}
}

\begin{document}
\pagestyle{empty}
\maketitle
\thispagestyle{empty}

\begin{abstract}
This paper focuses on self-supervised monocular depth estimation in dynamic scenes trained on monocular videos. Existing methods jointly estimate pixel-wise depth and motion, relying mainly on an image reconstruction loss. Dynamic regions\footnote{Dynamic regions indicate regions covered by moving objects.} remain a critical challenge for these methods due to the inherent ambiguity in depth and motion estimation, resulting in inaccurate depth estimation. This paper proposes a self-supervised training framework exploiting pseudo depth labels for dynamic regions from training data. The key contribution of our framework is to decouple depth estimation for static and dynamic regions of images in the training data. We start with an unsupervised depth estimation approach, which provides reliable depth estimates for static regions and motion cues for dynamic regions and allows us to extract moving object information at the instance level. In the next stage, we use an object network to estimate the depth of those moving objects assuming rigid motions. Then, we propose a new scale alignment module to address the scale ambiguity between estimated depths for static and dynamic regions. We can then use the depth labels generated to train an end-to-end depth estimation network and improve its performance.
Extensive experiments on the Cityscapes and KITTI datasets show that our self-training strategy consistently outperforms existing \term depth estimation methods. Our code is available at \small \url{https://github.com/HoangChuongNguyen/mono-consistent-depth.git}



\end{abstract}    

\section{Introduction}
\label{sec:introducntion}

\begin{figure}
    \scriptsize
    \centering
    \qquad  
    \setlength\tabcolsep{1.0pt}
    \begin{tabular}{c c c}
        \multirow{1}{*}[+6.0ex]{\rotatebox{90}{Image}} & \includegraphics[width=.225\textwidth]{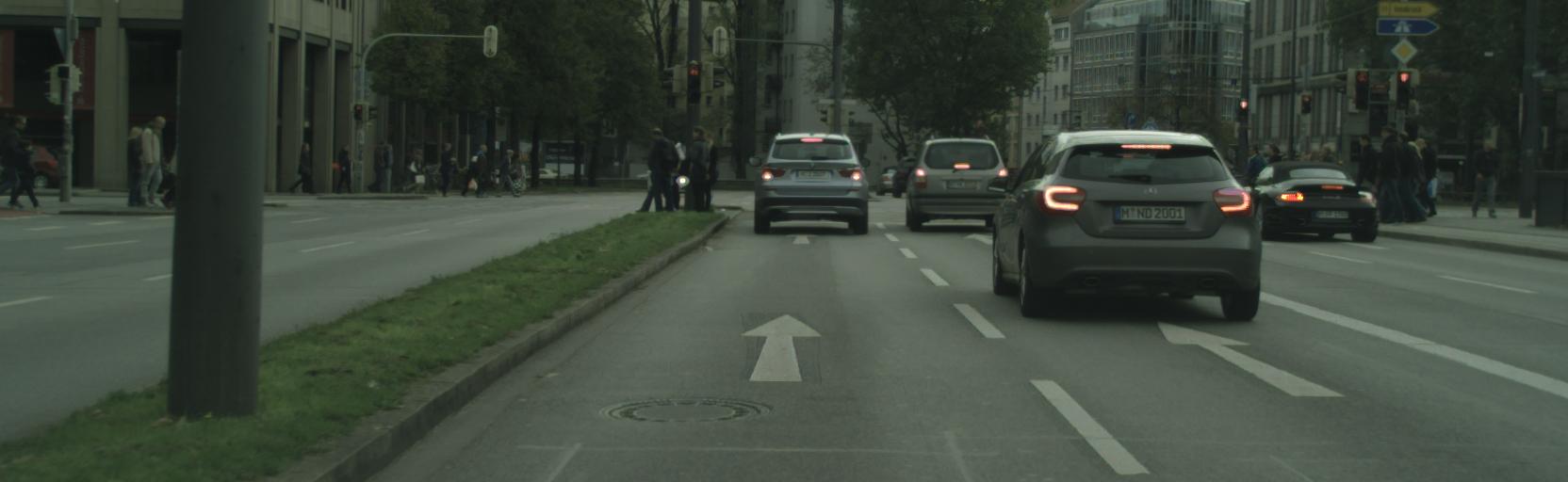} & \includegraphics[width=.225\textwidth]{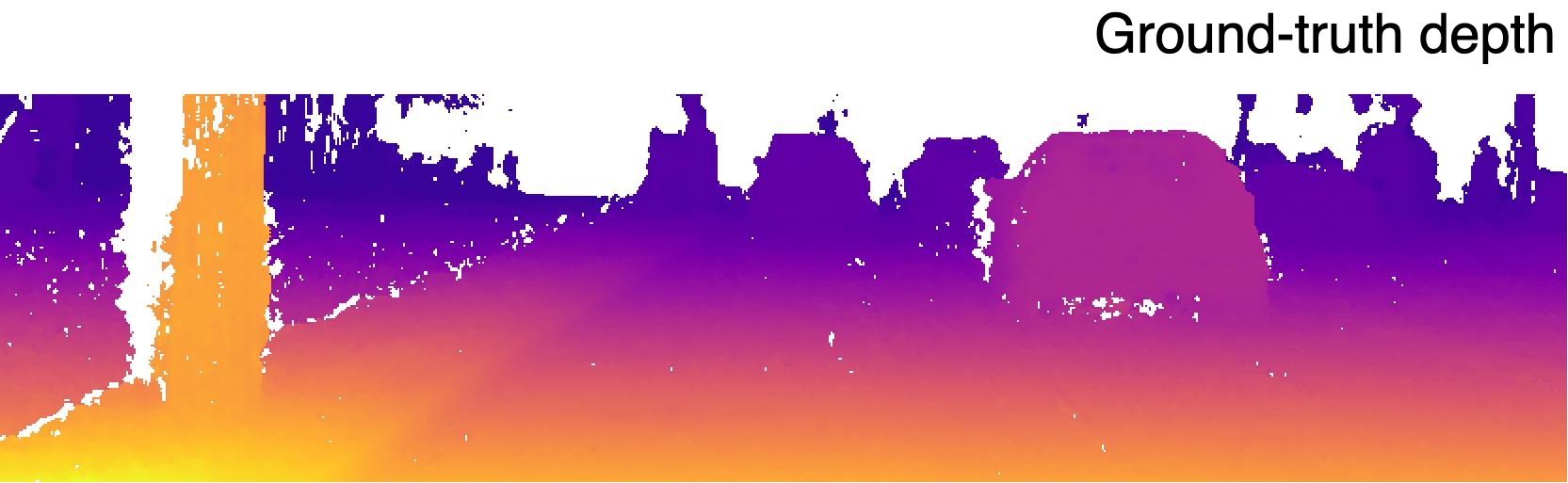} \\ 

        \multirow{1}{*}[+4.5ex]{\rotatebox{90}{ \cite{lee2021learning}}} & \includegraphics[width=.225\textwidth]{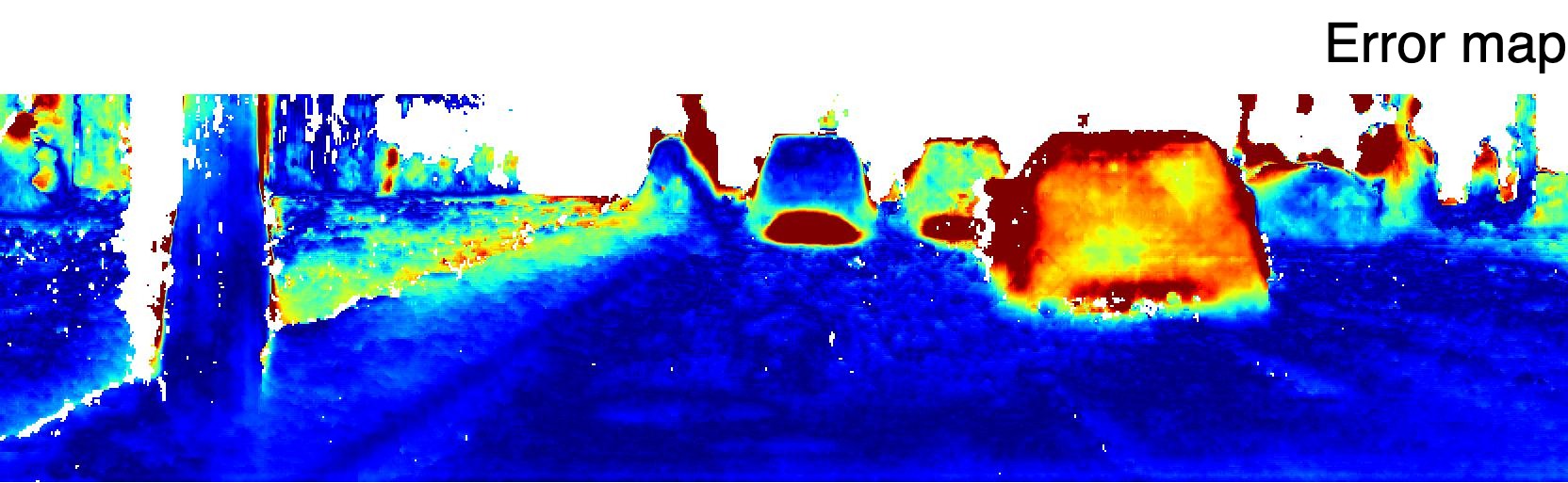} & \includegraphics[width=.225\textwidth]{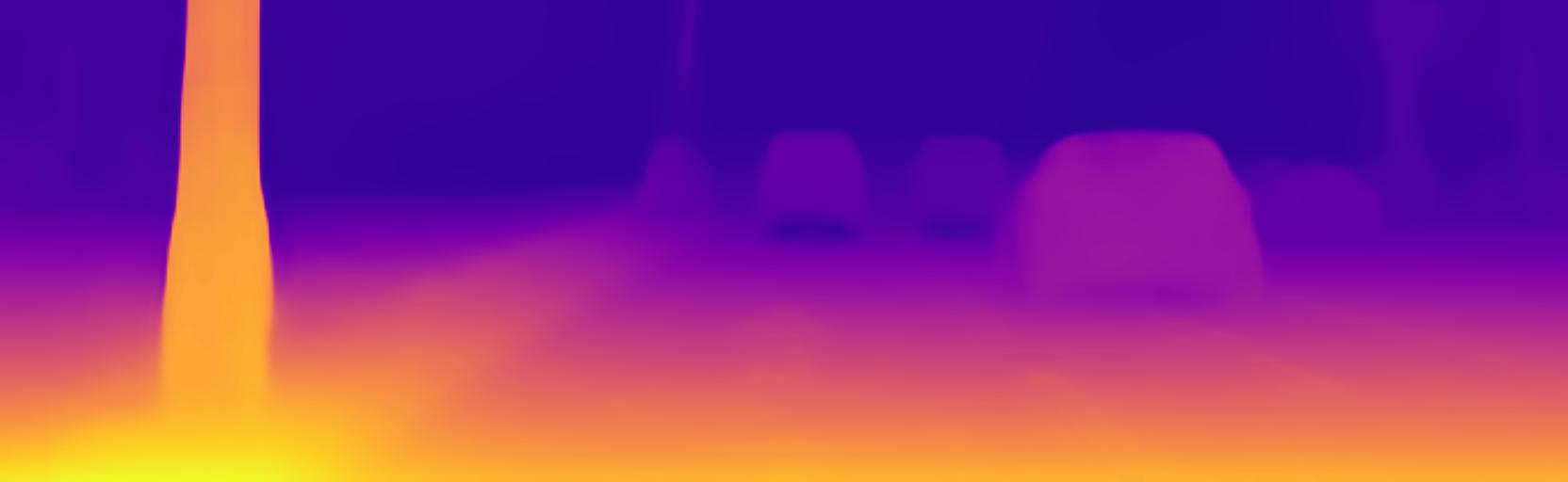} \\ 

        \multirow{1}{*}[5.0ex]{\rotatebox{90}{Ours}} & \includegraphics[width=.225\textwidth]{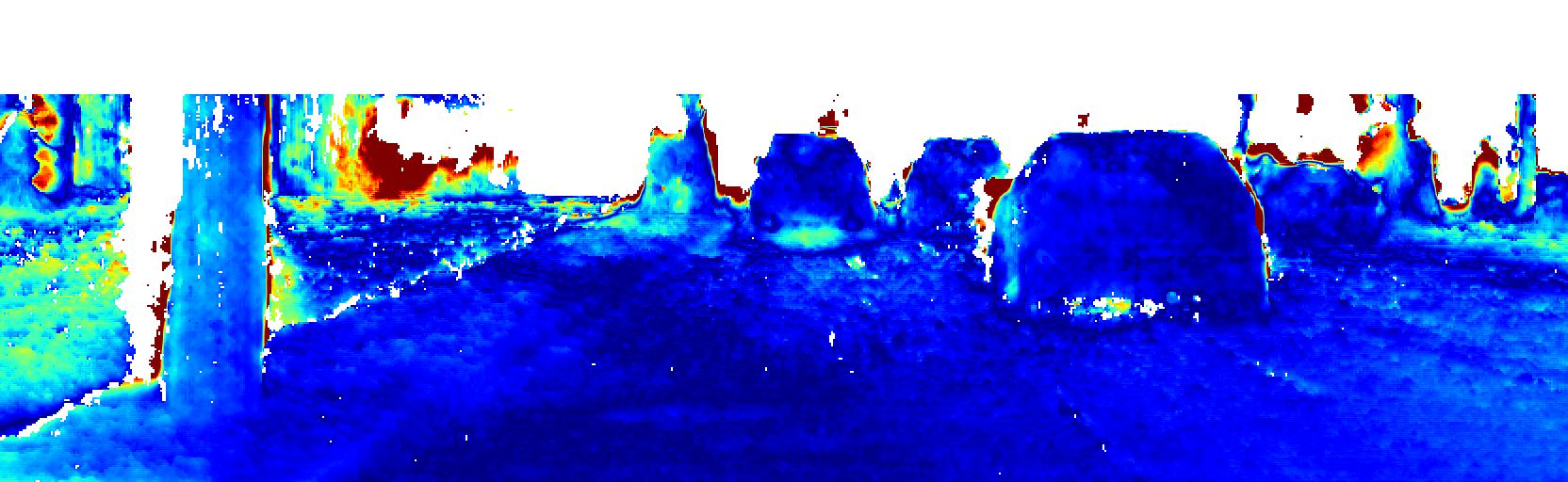} & \includegraphics[width=.225\textwidth]{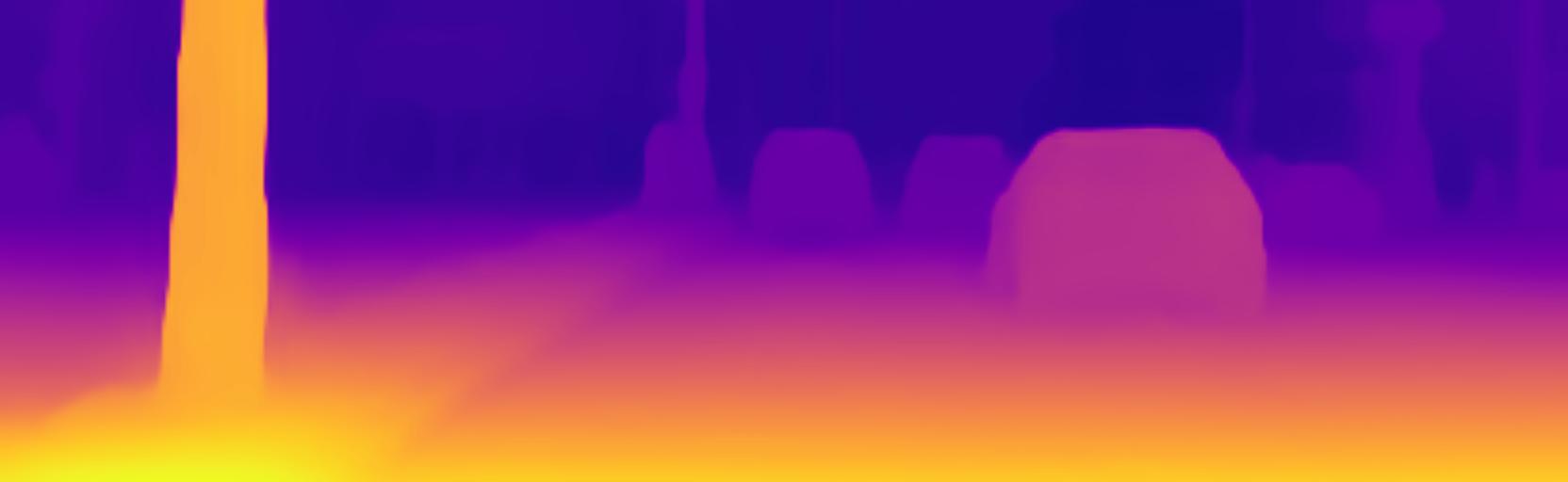}
   
    \end{tabular}
    \caption{Depth predictions for an image. Areas with more intense red color in the error maps represent higher error. Our method produces more accurate depths for moving objects in the image. }
    \label{tab:my_label} 
        
    
\end{figure}




Estimating a dense depth map from a monocular image is crucial for various applications, such as 3D reconstruction~\cite{sargent2023vq3d}, autonomous driving \cite{sun2020scalability}, and augmented reality \cite{zhang2021consistent}. Self-/unsupervised depth estimation using monocular video sequences is an effective way to bypass the need for expensive depth labels~\cite{casser2019depth, godard2019digging, guizilini20203d}. 
Current methods minimize the image reconstruction loss during training to jointly estimate depth and camera pose~\cite{godard2019digging, han2023self, han2022brnet, guizilini20203d}. These methods suffer accuracy degradation in the presence of moving objects, e.g, cars, as the reconstruction process does not take motion into consideration. To address this limitation, a few works infer the camera and pixel-wise object motion in 3D, and the depth to reduce the image reconstruction error~\cite{li2021unsupervised, hui2022rm}. Other approaches incorporate binary masks for moving objects to improve the pixel-wise motion estimation~\cite{casser2019depth,lee2021learning}. Despite the accuracy improvement, as we will show in Section~\ref{sec:scaleAM}, there are inherent ambiguities entangling motion and depth estimation in dynamic regions that can not be fully resolved using regularization terms or joint optimization, i.e., enforcing sparsity on pixel-wise motion. In contrast, we propose to exploit~\emph{scale-consistent} pseudo depth labels for moving objects which we can then use to train a monocular depth estimation network.

In contrast to existing methods, we propose to decouple the depth estimation for static and dynamic regions in images. We first leverage existing self-supervised depth estimation frameworks to jointly estimate the depth, pixel-wise object motion, and camera rigid motion~\cite{li2021unsupervised}. We use pixel-wise object motion to identify moving objects, e.g., the foreground, and their pixel-wise correspondences across frames. For the rest of the image, the background, we directly use the inferred depth map as depth labels. 

For the foreground, the goal is to obtain pseudo depth labels for each object that are scale-consistent with the background. We achieve this in two steps. First, we employ an Object Network to estimate the depth label for the dynamic regions for each frame in the sequence. This network takes a pair of images with corresponding objects as input to jointly estimate the relative camera motion, namely, rigid rotation and translation, and depth map for the object in the reference frame via the image reconstruction loss. This can be viewed as a dense structure-from-motion process that leads to scaled depth and camera pose estimation. Second, we propose a scale alignment network to resolve scale ambiguities between each object in the foreground and the background. We then obtain scale-consistent depth pseudo labels for the foreground that can be used as additional supervision signals for training depth estimation networks. 

We demonstrate the benefits of using depth pseudo labels obtained using our approach on several existing Self-/unsupervised monocular depth estimation frameworks on the KITTI and Cityscapes datasets. Our results show consistent overall improvements over frameworks and datasets with significant error reduction on dynamic regions.

In summary, our contributions are:
\begin{itemize}
 \item We introduce a new scale-alignment network to solve scale-ambiguities among objects and the background.
    \item To the best of our knowledge, we are the first to extract scale-consistent pseudo depth labels as supervision signal to solve the scale-ambiguity problem commonly present in monocular depth estimation for dynamic scenes.
    
    \item Our approach consistently outperforms all previous \term depth estimation methods by large margins for dynamic regions. Compared to prior works, we achieve up to $\mathbf{52.6\%}$ and $\mathbf{14.4\%}$ error reduction on the Cityscapes and KITTI datasets, respectively. 
    
\end{itemize}

\section{Related Work}
\noindent {\bf \Term Learning from Stereo Images.}
To avoid costly ground-truth data for training a depth network, \cite{garg2016unsupervised} introduces an approach under the unsupervised framework for depth estimation using a synchronized stereo pair as training data. \cite{godard2017unsupervised} imposes a consistency constraint on the predicted disparity when a stereo pair is adopted as a supervision signal. Generative adversarial networks are also explored for this task \cite{aleotti2018generative, pilzer2018unsupervised}. \cite{yang2020d3vo} attempts to improve the training signal by aligning the lighting condition of the input images and modeling pixel photometric uncertainties. Recently, \cite{wang2023planedepth} proposed to estimate a depth distribution based on the concept of orthogonal planes in the 3D world. In this work, we show that we can exploit reliable depth supervision signals from monocular video instead of stereo pairs for practical applications.

\noindent{\bf \Term Learning from Monocular Video.}
The monocular video provides a less constrained form of \term learning. \cite{zhou2017unsupervised} firstly introduces a promising framework that jointly learns depth and camera ego-motion from consecutive frames in a video. Later on, \cite{godard2019digging} improves the training signal by introducing the per-pixel minimum reprojection loss to filter out occluded pixels from the final image reconstruction loss. More robust geometric constraints enforcing consistency between predicted point clouds or predicted depth across consecutive frames are proposed by \cite{mahjourian2018unsupervised, bian2019unsupervised}. To further boost the performance, several works propose new architectures for a depth estimation network \cite{guizilini20203d, zhou2021self, han2022brnet, han2023self}. More recently, cost volume-based approaches that predict depth from multiple images are introduced \cite{watson2021temporal, chen2023self, guizilini2022multi}. 
Although significant improvements have been achieved, all the mentioned methods tend to have inferior performance in dynamic regions. The main challenge arises from the presence of moving objects in the scene, which does not satisfy the photometric consistency assumption given no specific consideration for object motion.  

To address the issues caused by moving objects, \cite{casser2019depth} proposed a joint modelling of depth, camera motion, and object motion. 
Specifically, a warped image is first obtained by relying on the initially predicted depth and camera motion. 
This warped image and its corresponding reference image are then used to predict object rigid motion. 
\cite{lee2021learning} improves this framework by applying forward warping to encourage consistent geometry while modelling object motion. One disadvantage of such methods stems from the ambiguity of jointly predicting object depth and object motion. 
In particular, incorrect initial depth prediction could lead to incorrect warped images, which in turn causes errors in the predicted object motion. Nonetheless, there exist multiple combinations of depth and object motion that result in the minimization of the image reconstruction loss, namely the inherent ambiguities due to insufficient constraints \cite{casser2019depth,lee2021learning}. 
In lieu of predicting object rigid motion, \cite{li2021unsupervised} proposes to predict pixel-wise motion. 
Inspired by previous work, \cite{hui2022rm, lee2021attentive} introduces a two-stage motion disentanglement, which is similar to \cite{casser2019depth} but with pixel-wise motion prediction. 
Nevertheless, these approaches exacerbate the scale ambiguity issue as there might be variations in the scale factor associated with different pixels of the same moving object. In this work, we propose to estimate high-quality scale-consistent depth pseudo labels for the moving objects and static background in separate stages which are used further to improve depth predictions for dynamic regions.

\vspace{-1.0mm}
\section{Preliminary}\label{sec:scaleAM}
\vspace{-0.5mm}
\begin{figure*}
    \centering
    \includegraphics[width=\textwidth]{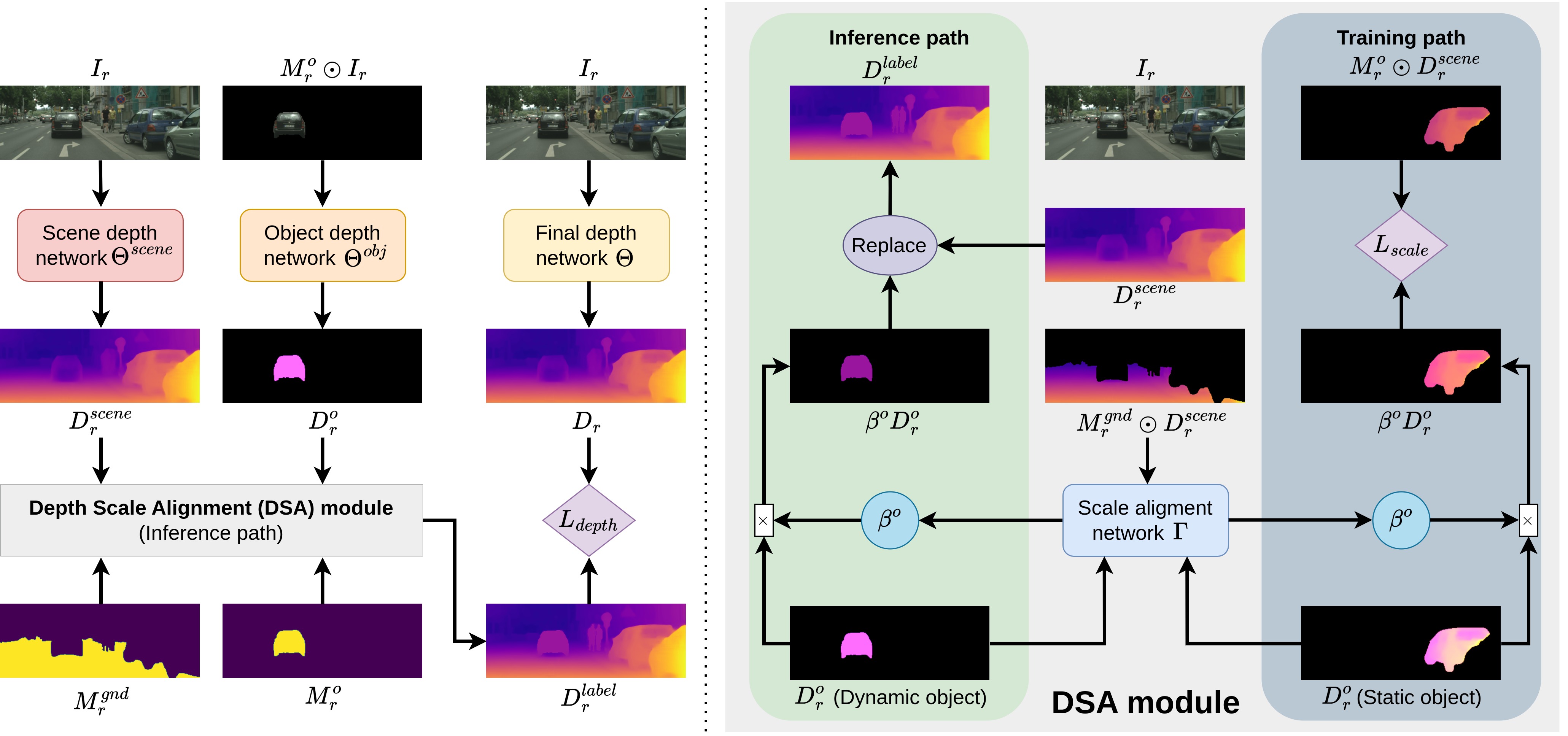}
    \caption{Our proposed framework. \textbf{Left}: How the pseudo depth label $\mathbf{D}_r^{label}$ is extracted and used to train the final depth network. \textbf{Right}: How the DSA module is trained (training path) and utilized to produce $\mathbf{D}_r^{label}$ (inference path). The DSA is trained to predict a scale $\beta^o$ that minimizes the difference between two depth predictions of the same static object. It is then used to align the depth scale of every dynamic object to the depth scale of the static region. $\mathbf{D}_r^{label}$ is constructed from depth of static regions in $\mathbf{D}_r^{scene}$ and depth of dynamic objects after alignment $\beta^o \mathbf{D}_r^o$.}
    \label{fig:3_method}
    \vspace{-5mm}
\end{figure*}

\noindent{\bf Scale-consistent Depth Estimation.}
Scale ambiguity is a phenomenon often encountered in \term dynamic scene monocular depth estimation.
This setup involves two consecutive RGB frames, termed as reference and source frame $\mathbf{I}_r, \mathbf{I}_s \in \mathbb{R}^{H \times W \times 3}$, from a monocular video with known camera intrinsic matrix $\mathbf{K}$ but unknown poses. 
For a static scene, the correspondence between pixels in these frames can be established via
\begin{equation}
     \mathbf{x}' d' = \mathbf{K} (\mathbf{R} \mathbf{K^{-1}} \mathbf{x}d + \mathbf{T}),
\end{equation}
where $\mathbf{x}$ and $\mathbf{x}'$ is the homogeneous coordinate of a pixel in $\mathbf{I}_r$ and $\mathbf{I}_s$ respectively.  $d$ and $d'$ represent the depth values for ${\bf x}$ and ${\bf x}'$ in their respective frames. $\mathbf{R}$, $\mathbf{T}$ specify (inferred) relative camera pose that is shared across all pixels.
However, this mapping is not unique.
One can scale both sides of the equation with a constant $c$ for \textbf{all pixels} at the same time without violating the equality. That is, 
\begin{equation}
    \mathbf{x}'(cd') = \mathbf{K} (\mathbf{R} \mathbf{K^{-1}} \mathbf{x} (cd )
 + c\mathbf{T}).
\end{equation}
Thus, the estimated depth $d$ for all pixels is at best one unified scale away from the ground truth as the scale $c$ can not be uniquely determined even with the correct correspondence.
We say such depth prediction with a unified scale shared across all pixels is \textit{scale-consistent}.

However, the presence of moving objects violates such scale consistency.
To capture object motions, we need to introduce a translation vector $\mathbf{t}_{x}$ that can vary from pixel to pixel and establish the correspondence as
\begin{equation}
    \mathbf{x}'d' = \mathbf{K} (\mathbf{R} \mathbf{K^{-1}} \mathbf{x}d + \mathbf{T}) + \mathbf{t}_{x}
\end{equation}
In this case, one can scale the depth of the pixel via an arbitrary constant $c$ and always find a $\mathbf{t}_x$ to retain the correspondence via
\begin{equation}
     \mathbf{t}_{x} = d' \mathbf{x}' - \mathbf{K} (\mathbf{R} \mathbf{K^{-1}} \mathbf{x}(cd) + \mathbf{T}).
\end{equation}
Thus, two pixels undergoing different motions can have different depth scales, which will lead to depth values with arbitrary differences.
We refer to this phenomenon as \textit{scale ambiguity} and term a depth prediction without unified scale as \textit{scale-inconsistent}.

\vspace{-1.0mm}
\section{Method}
\vspace{-0.5mm}

Our goal is to estimate the depth map ${\bf D}_r$ from a monocular image ${\bf I}_r$, namely ${\bf D}_r=\Theta({\bf I}_r)$ where $\Theta$ denotes the depth estimation network. In particular, we aim to generate high-quality~\emph{scale-consistent} pseudo depth labels ${\bf D}_r^{label}$ from a monocular video for self-supervised learning of ${\Theta}$. To this end, we propose to estimate a decoupled representation 
\begin{equation}\label{eq:dlabel}
   {\bf D}_r^{label} = (1-\mathbf{M}_r)\odot{\bf D}_r^{scene} + \mathbf{M}_r \odot{\bf D}^{dyn}_r, 
\end{equation} where ${\bf D}_r^{scene}$, and ${\bf D}^{dyn}_r$, and $\mathbf{M}_r$ represent depth label for the background scene, the moving objects, and moving object masks, respectively. An overview of our framework is shown in Fig. \ref{fig:3_method}. In the following, we introduce the modules used to generate masks and depth labels employed above.

\vspace{-0.5mm}
\subsection{Scene Depth Estimation for ${\bf D}_r^{scene}$}
\label{section:scene_depth_estimation}



In this stage, we aim to obtain the scene depth ${\bf D}_r^{scene}$ for a reference image ${\bf I}_r$ through unsupervised learning. 
At training time, given $\mathbf{I}_r, \mathbf{I}_s \in \mathbb{R}^{H \times W \times 3}$, from a monocular video with known $\mathbf{K}$, we formulate the problem as the minimization of image reconstruction error using depth ($\mathbf{D}^{scene}_r$) and camera pose ($ \mathbf{R}^{cam}, \mathbf{T}^{cam}$) as intermediate variables. To handle moving objects, following~\cite{li2021unsupervised}, we further introduce the pixel-wise object motion $\Delta_r$ as an additional intermediate variable to generate a synthesized image $\mathbf{I}_{r \leftarrow s}$.



Instead of predicting these intermediate variables in parallel, we first adopt a depth network $\Theta^{scene}$ and a camera pose network $\Phi^{cam}$, following the pipeline design in \cite{li2021unsupervised, lee2021learning},  to predicate 
$\mathbf{D}^{scene}_r \in \mathbb{R}^{H \times W \times 1}$ and 
$\mathbf{R}^{cam}$, $\mathbf{T}^{cam}$  by 
$        \mathbf{D}^{scene}_r = \Theta^{scene}(\mathbf{I}_r); \;
        \mathbf{R}^{cam}, \mathbf{T}^{cam} = \Phi^{cam}(\mathbf{I}_r,\mathbf{I}_s).$
Given the camera relative pose, a pixel $\mathbf{x}$ in the reference frame with depth $d = \mathbf{D}_{r}^{scene}(\mathbf{x})$ is mapped to $\mathbf{x'}$ with depth $d'$ in the source frame via the bijection $\tau$ defined as
\begin{equation}
    \mathbf{x'}d' = \tau(\mathbf{x}d) := \mathbf{K}(\mathbf{R}^{cam}\mathbf{K}^{-1} \mathbf{x}d + \mathbf{T}^{cam}).
\label{equation:rigid_motion_mapping}
\end{equation}
Given $\tau$, we can compute a reconstruction of the source view $\mathbf{I}_{r \rightarrow s}$ from the reference view $\mathbf{I}_r$ via  
$    \mathbf{I}_{r \rightarrow s} = \mathcal{FW}(\tau, \mathbf{I}_r),$
where $\mathcal{FW}$ denotes the forward warping operation.
We then employ a network $\Psi$ to predict a motion map $\Delta_r \in \mathbb{R}^{H \times W \times 3}$ taking $\hat{\mathbf{I}}_{r \rightarrow s}$ and $\mathbf{I}_s$ as input,
\begin{equation}
    \Delta_r = \mathcal{FW}(\tau^{-1}, \Psi(\mathbf{I}_{r \rightarrow s}, \mathbf{I}_s)).
\end{equation}
We get a refined correspondence  $\tau'$ defined as
\begin{equation}
    \mathbf{x''}d'' = \tau'(\mathbf{x}d) := \tau(\mathbf{x}d) + \Delta_r(\mathbf{x}).
    \label{equation:3_find_full_correspondences}
\end{equation}
The reference image reconstruction $\mathbf{I}_{r \leftarrow s}$ is induced by $\tau'$ via
\begin{equation}
    \mathbf{I}_{r \leftarrow s} = \mathcal{BW}(\tau', \mathbf{I}_{s}),
    \label{equation:bw_warp_final_image}
\end{equation}
\noindent where $\mathcal{BW}$ the backward warping operation.

The loss function for training the networks above includes the reference image reconstruction loss $L_p$ \cite{godard2019digging}, depth edge-aware smoothness loss $L_s$ \cite{godard2017unsupervised}, and object-motion sparsity loss $L_g$ \cite{li2021unsupervised}:
\begin{equation}
    L_{scene} = L_p(\mathbf{I}_{r\leftarrow{s}},\mathbf{I}_r) + L_s(\mathbf{D}^{scene}_r, \mathbf{I}_r) + L_g(\Delta_r),
    \label{equation:loss_function_stage1}
\end{equation}
where $L_g(\cdot)$ encourages the predicted pixel-wise motion to be zero at static background regions in the scene.
As discussed in~Sec.~\ref{sec:scaleAM}, the static regions of $\mathbf{D}^{scene}_r$ are scale-consistent. 

\subsection{Dynamic Object Depth Estimation for ${\bf D}^{label}_r$}
\label{section:object_depth_estimation}
At this stage, we focus on the scale-consistent depth estimation for moving object regions. As discussed in the previous section, we can obtain a pixel-wise motion map that leads to the estimation of the object masks (see Sec.~\ref{section:mask_segmentation} for details).  Given the objects' masks and images of the object region only, we first estimate an initial object depth via unsupervised learning by minimizing the image reconstruction loss. Note that the estimated depth for moving objects is not scale consistent with the static regions. We therefore introduce a scale-alignment network to resolve the scale ambiguities. Below we provide details for these two steps.

\noindent{\bf Self-supervised Object Depth Estimation.} Similar to the previous section, we treat depth and camera pose as the intermediate variables for unsupervised learning. Given the mask for object $o$ in the reference view as $\mathbf{M}^o_r  \in \{0, 1\}^{H \times W \times 1}$, we can map it to the source view and obtain the corresponding object mask as
$\mathbf{M}_{r\rightarrow{s}}^{o} = \mathcal{FW}(\tau', \mathbf{M}^{o}_{r}).$

We then estimate the depth and motion of the object $o$ as
\begin{equation}
    \mathbf{D}_r^{o} = \Theta^{obj}(\mathbf{M}^{o}_r \odot \mathbf{I}_r),
\end{equation}
\begin{equation}
    \mathbf{R}^{o}, \mathbf{T}^{o} = \Phi^{obj} (\mathbf{M}^{o}_r\odot \mathbf{I}_r,\mathbf{M}_{r\rightarrow{s}}^{o}\odot \mathbf{I}_s).
\end{equation}
$\mathbf{R}^{o}$ and $\mathbf{T}^{o}$ represent view changes caused by both object rigid motion and camera motion, $\odot$ defines the element-wise multiplication operation. By substituting $\mathbf{D}^{o}_r$, $\mathbf{R}^{o}$, and $\mathbf{T}^{o}$ into Eq. \ref{equation:rigid_motion_mapping}, we can solve for pixel correspondence $\tau^{o}$ and synthesize the masked reference image $\mathbf{I}_{r\leftarrow{s}}^{o}$ via
\begin{equation}
    \mathbf{I}^o_{r \leftarrow s} = \mathcal{BW}(\tau^o, \mathbf{M}^o_{r \rightarrow s} \odot \mathbf{I}_{s}).
\end{equation}
 The loss function used to train $\Theta^{obj}$ and $\Phi^{obj}$ is, 
\begin{equation}
    L_{obj} = L_p(\mathbf{I}^o_{r \leftarrow s},\mathbf{M}^{o}_r\odot \mathbf{I}_r) + L_s(\mathbf{M}^{o}_r \odot \mathbf{D}^{o}_r,\mathbf{M}^{o}_r).
\end{equation}

\noindent{\bf Depth Scale Alignment}.
In the previous stage, we unified the depth scale within each object. 
However, the scales still mismatch between different moving objects and static regions in $\mathbf{D}^{scene}_r$. To align the scales, we introduce the Depth Scale Aligment (DSA) module. The right part of Fig. \ref{fig:3_method} depicts the training process and how this module is utilized to address the scale ambiguity issue for moving objects. 

The DSA module is designed based on three key observations.
 \textbf{Obs.\:1:} \label{obs:1} In $\mathbf{D}^{scene}_r$, depth values of pixels within static objects exhibit a unified scale while moving objects still exhibit pixel-level scale-inconsistency due to the pixel-wise motion. By contrast, $\Theta^{obj}$ can predict depth values for all the pixels at a unified scale within each static as well as the moving object. Thus, for static object only, the depth values predicated by $\Theta^{scene}$ and $\Theta^{obj}$ only differ by one single scale whereas it does not hold for moving objects.  \textbf{Obs.\:2:} The scale-consistent depths of objects should have the similar range as the those of static regions they are situated on or close to. Thus, the objects' depth scale ambiguity could be resolved based on the information from the depth value of its nearby static regions.
\textbf{Obs.\:3}: Moving objects are indistinguishable from static objects when only one frame is observed, as the motion information is lost. Thus, a single frame model trained to predict the scale ratio between the inferred depth from $\Theta^{scene}$ and $\Theta^{obj}$ for static object can be generalized to moving objects after training.


The above observations indicate that we are able to train a DSA network $\Gamma$ to predict a scale alignment ratio $\beta \in \mathbb{R}^+$ to align the depth of a static object inferred by $\Theta^{obj}$ to its desired depth predicted by $\Theta^{scene}$. In other words, ground truth of the aligned depth scale for static objects are known. We thus only train the network $\Gamma$ relying on static objects. 
Below we describe the training of $\Gamma$. 

We denote the set of static objects as $O^s$ and moving objects as $O^d$ with $O = O^s \cup O^d$ which are obtained based on the segmentation method described in~\ref{section:mask_segmentation}.
As stated in Obs.\:1, we only consider \textbf{static} objects $O^s$ in training. For each static object $o \in O^s$, as suggested by Obs.\:2, $\Gamma$ takes $\mathbf{I}_r$, $D_r^{o}$, and ground depth to infer scale alignment ratio $\beta^o$:
\begin{equation}
    \beta^o = \Gamma(\mathbf{I}_r, \mathbf{D}_r^{o}, \mathbf{M}^{gnd}_r \odot \mathbf{D}_r^{scene}).
\end{equation}
The loss function computes the difference between the depth predicted by $\Theta^{obj}$ and $\Theta^{scene}$ but after alignment as
\begin{equation}
    L_{scale} = \frac{\sum_{u,v} \lvert\beta^o \mathbf{D}^{o}_r(u, v) - \mathbf{M}^{o}_r(u, v) \mathbf{D}^{scene}_r (u, v)\rvert}    {\sum_{u, v} \mathbf{M}^{o}_r(u, v)},
\end{equation}
where the scale alignment is done by multiplying depth prediction $\mathbf{D}_r^{o}(u, v)$ by $\beta^{o}$.

After training, Obs.\:3 indicates that we can perform the inference of the DSA network on \textbf{moving} objects and obtain the depth ratio between each \textbf{moving} object and the static regions to achieve a dynamic region depth prediction
$\mathbf{D}^{dyn}_r = \sum_{o \in O^d} \beta^o \mathbf{M}^{o}_r \odot 
 \mathbf{D}^{o}_{r}$ that share the same scale with the static regions within the same image.
 We define the dynamic region mask as $\mathbf{M}_{r} = \sum_{o\in O^{d}} M_r^o$.
Then, an image depth prediction as the depth label ${\bf D}_r^{label}$ with a unified scale is generated based on Eq.~\ref{eq:dlabel},
The depth scale alignment component is the key to achieving scale-consistent full image depth prediction and improves depth prediction accuracy by a large margin as demonstrated in the experiment section below.

\subsection{Mask Generation \textbf{${\bf M}^o_r, {\bf M}^{gnd}_r$}}
\label{section:mask_segmentation}


The masks for objects and the ground in the image are key components to obtain the scale-consistent depth estimation. In principle, we can adopt existing instance segmentation methods, a pre-trained segmentation model such as~\emph{SEEM}~\cite{zou2023segment} for objects and ground which leverages annotations, or build a self-supervised pipeline using existing unsupervised approaches to discover the object and ground masks. Below, we introduce self-supervised pipelines that can discover object masks and ground masks, respectively, to demonstrate that we can derive the segmentation masks in a self-supervised manner.

\noindent\textbf{Self-supervised Object Detection.} We adopt the slot-attention (SA) model \cite{locatello2020object} to obtain the pseudo moving object masks from the object motion map and depth map we obtained in~Section~\ref{section:scene_depth_estimation}. Given the extracted dynamic object masks, we could train a~\emph{MaskRCNN} using these object masks to discover all the object regions (${\bf M}_r^o$) including static ones which are required in the scale-alignment process. Details are provided in the Supplementary material.

\begin{figure*}
    \centering
    \begin{subfigure}{0.33\textwidth}
        \includegraphics[width=\linewidth]{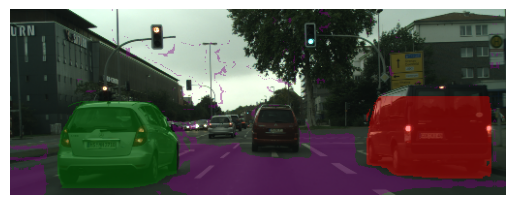}
        \vspace{-5.0mm}
        \caption{Our pseudo mask}
    \end{subfigure}
    \hfill
    \begin{subfigure}{0.33\textwidth}
        \includegraphics[width=\linewidth]{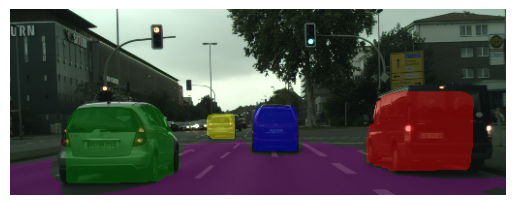}
        \vspace{-5.0mm}
        \caption{Our self-supervised segmentation}
    \end{subfigure}
    \hfill
    \begin{subfigure}{0.33\textwidth}
        \includegraphics[width=\linewidth]{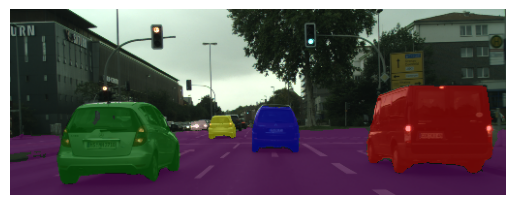}
        \vspace{-5.0mm}
        \caption{SEEM segmentation}
    \end{subfigure}
    \vspace{-6.0mm}
    \caption{Qualitative segmentation results. Static cars (two in the middle, stopped by the traffic light) originally ignored by the pseudo label are successfully segmented out by both our self-supervised model as well as SEEM. }
    \label{fig:3.5_masks_visualization}
\end{figure*}
\vspace{-5.0mm}

\noindent\textbf{Self-supervised Ground Segmentation.}
To obtain ground pseudo masks, we follow the method introduced by \cite{xue2020toward} and segment ground pixels according to their height in the 3D world and surface normal. Then, we train a segmentation model to predict the pseudo ground mask $\mathbf{M}^{gnd}_r$ with the binary-cross entropy loss.
We show in Fig.~\ref{fig:3.5_masks_visualization} a qualitative comparison between the pseudo label, self-supervised segmentation, and \emph{SEEM} segmentation results. Compared with the pseudo label, the self-supervised model is able to discover both static and dynamic moving objects with similar results using \emph{SEEM}. Please see supplementary for more details.

\subsection{Final Depth Prediction ${\bf{D}}_{r}$}
\label{section:final_depth_prediction}

Getting the scale-consistent depth prediction $\mathbf{D}^{label}_{r}$ involves the evaluation of multiple networks.  We propose to train a depth network  $\Theta$ with $D^{label}_r$  to directly predict scale-consistent depth from an RGB image in one go,
\begin{equation}
    \mathbf{D}_{r} = \Theta(\mathbf{I}_r).
\end{equation}

Although the loss function used to train this model can be as simple as a regression loss between its predictions and the pseudo label $\mathbf{D}_r^{label}$, we find that loss terms employed in the first stage prevent the model from learning the inaccurate pseudo depth label (e.g., induced by inaccurate object masks).  The loss used to train the models is                      
\begin{equation}
\begin{split}
    L_{final} &= L_p(\mathbf{I}_{r\leftarrow{s}},\,\mathbf{I}_r) + L_s(\mathbf{D}_{r},\mathbf{I}_r) \\
    &+ L_g(\Delta_{r}) + L_{depth}(\mathbf{D}_{r}, \mathbf{D}_{r}^{label}, \mathbf{M}_r).
    \label{eq:final}
\end{split}
\end{equation}
The first three terms in Eq. \ref{eq:final} are identical to $L_{scene}$ with
$L_{depth}$ provides scale-consistent dynamic region supervision and is defined as

\begin{equation}
    L_{depth} = \frac{\sum_{u,v} \mathbf{M}_{r}(u,v) \lvert \mathbf{D}_{r}(u, v) - \alpha \mathbf{D}^{label}_r (u, v)\rvert}    {\sum_{u, v} \mathbf{M}_{r}(u,v)},
\end{equation}
\begin{equation}
    \alpha = \text{Median} \left( \left\{\frac{\mathbf{D}_{r}(u,v)}{\mathbf{D}^{label}_r(u,v)} \right\}_{u,v} \right),
\end{equation}
where $\alpha$ is the median of the per-pixel depth ratio between $D_r$ and $D_r^{label}$ used to align the scale of $D_r^{label}$ to $D_r$, effectively making $L_{depth}$ scale-invariant. The rationale behind the design is that $D_r^{label}$, as a scale-consistent pseudo label, is itself one scale away from the ground truth depth. A direct regression without $\alpha$ will force our network to predict depth at the specific scale. However, as the scale itself is arbitrary, there is no reason to prefer one scale over the other. By applying $\alpha$ we allow the network to find its own scale without introducing scale-inconsistency, lightening the learning burden.

We initialize $\Theta$ with $\Theta^{scene}$, i.e,  instead of training from scratch, we fine-tune $\Theta^{scene}$.
The camera pose network  $\Phi^{cam}$ and the pixel-wise motion network $\Psi$ are also fine-tuned at this stage.

\vspace{-2.0mm}
\section{Experiments}
\vspace{-1.0mm}
\subsection{Implementation Details}
\vspace{-1.0mm}

\noindent\textbf{Dataset and Evaluation Metrics} We evaluate our method on the Cityscapes and KITTI datasets. For Cityscapes, we use 69,731 images for training and 1,525 images for testing. For KITTI, we use 39,810 training images and evaluate our method on 697 images from the Eigen split \cite{eigen2015predicting}. Additionally, we manually annotate moving-object masks for test set images for both datasets. The annotation allows us to evaluate performance on static and dynamic regions separately.

Tab.~\ref{Tab:moving_object_stats} reports the statistics on the number of dynamic objects. The table reveals that the KITTI dataset primarily consists of static scenes, making it less prone to scale ambiguity. On the other hand, the Cityscapes dataset features a large amount of moving objects. For this reason, our experiments mainly focus on the Cityscapes dataset. Regarding evaluation metrics, we adopt the absolute relative error (Abs Rel), square relative error (Sq Rel) and the accuracy metric for depth prediction ($\delta<1.25$) as used in \cite{godard2019digging}. 

\vspace{-2.0mm}
\begin{table}[h]
    \small
    \centering
    \begin{tabular}{|c|c|c|}
         \hline
         Dataset&  Dynamic objects/image & \% Dynamic pixels\\
         \hline
         KITTI&  0.828 & 0.6\% \\
         Cityscapes &  2.266 & 2.9\% \\
         \hline
    \end{tabular}
    \vspace{-2.0mm}
    \caption{Average number of dynamic objects and percentage of dynamic pixels in the Cityscapes and KITTI dataset.}
    \label{Tab:moving_object_stats}
\end{table}




\vspace{-2.0mm}
\noindent\textbf{Training and Network Details}
We perform the same data augmentation strategy as that is in \cite{watson2021temporal}. For our camera and object pose networks, we adopt the network structure from \cite{godard2019digging}.  We use \emph{RAFT} \cite{teed2020raft} as our pixel-wise motion network. 
The DSA modules consist of a \emph{ResNet18} \cite{he2016deep} backbone followed by 3 fully connected (FC) layers. \emph{ReLU} functions are applied after each FC layer except for the last one.
We explore several depth network architectures in our experiments. For each stage, we train models for 30 epochs with Adam optimizer \cite{kingma2014adam} with the learning rate set to 0.0002,  halved every 15 epochs. Exact forms of the loss function $L_{p}, L_{s}, L_{g}$ can be found in the supplementary material.

\vspace{-1.0mm}
\subsection{Results}
\vspace{-0.5mm}

\begin{table*}[h]
\footnotesize
\centering
\label{Tab:result_stage_wise_training}
\begin{tabular}{|c|c|c|c|ccc|ccc|ccc|}
\hline &  &   &   & \multicolumn{3}{c|}{Dynamic region}   & \multicolumn{3}{c|}{Static region}                              & \multicolumn{3}{c|}{All region}                                                         \\ \cline{5-13}
\multirow{-2}{*}{Backbone} & \multirow{-2}{*}{\begin{tabular}[c]{@{}c@{}}Pixel \\ motion\end{tabular}} & \multirow{-2}{*}{\begin{tabular}[c]{@{}c@{}}Pseudo\\ depth\end{tabular}} & \multirow{-2}{*}{\begin{tabular}[c]{@{}c@{}}Sematic\\ prior\end{tabular}} & \multicolumn{1}{c|}{Abs Rel} & \multicolumn{1}{c|}{Sq Rel} & $\delta < 1.25$ & \multicolumn{1}{c|}{Abs Rel} & \multicolumn{1}{c|}{Sq Rel} & $\delta < 1.25$ & \multicolumn{1}{c|}{Abs Rel} & \multicolumn{1}{c|}{Sq Rel} & $\delta < 1.25$ \\ \hline
\hline


\multirow{4}{*}{Resnet18 \cite{gordon2019depth}} &   &   &   & 0.352 & 11.169 & 0.858 & 0.118 & 1.362 & 0.862 & 0.127 & 1.743 & 0.722 \\

& \checkmark &   &   & 0.144 & 1.650 & 0.898 & 0.095 & 0.986 & 0.900 & 0.097 & 0.982 & 0.854 \\

& \checkmark & \checkmark &   & \underline{0.107} & \underline{0.784} & \textbf{0.906} & \underline{0.093} & \underline{0.961} & \textbf{0.907} & \underline{0.094} & \underline{0.931} & \underline{0.907} \\

& \checkmark & \checkmark & \checkmark & \textbf{0.100} & \textbf{0.647} & \underline{0.903} & \textbf{0.091} & \textbf{0.931} & \underline{0.904} & \textbf{0.092} & \textbf{0.895} & \textbf{0.911} \\
\hline
\hline
\multirow{4}{*}{Packnet \cite{guizilini20203d}}
&   &   &   & 0.182 & 2.667 & 0.891 & 0.104 & 1.167 & 0.894 & 0.108 & 1.210 & 0.813 \\
& \checkmark &   &   & 0.132 & 0.987 & \underline{0.906} & 0.091 & 0.861 & \underline{0.908} & 0.093 & 0.844 & 0.857 \\
& \checkmark & \checkmark &   & \underline{0.112} & \underline{0.703} & \textbf{0.908} & \underline{0.090} & \textbf{0.858} & \textbf{0.909} & \underline{0.091} & \underline{0.831} & \underline{0.890} \\
& \checkmark & \checkmark & \checkmark & \textbf{0.104} & \textbf{0.666} & \textbf{0.908} & \textbf{0.089} & \underline{0.859} & \underline{0.908} & \textbf{0.090} & \textbf{0.830} & \textbf{0.901} \\
\hline
\hline

 \multirow{4}{*}{DiffNet \cite{zhou2021self} } &   &   &   & 0.233 & 4.369 & 0.893 & 0.103 & 1.109 & 0.896 & 0.107 & 1.187 & 0.771 \\
& \checkmark &   &   & 0.152 & 1.646 & 0.912 & 0.087 & 0.831 & 0.917 & 0.090 & 0.833 & 0.828 \\
& \checkmark & \checkmark &   & \underline{0.113} & \textbf{0.657} & \underline{0.916} & \underline{0.083} & \textbf{0.775} & \underline{0.919} & \underline{0.085} & \textbf{0.753} & \underline{0.889} \\
& \checkmark & \checkmark & \checkmark & \textbf{0.105} & \underline{0.692} & \textbf{0.919} & \textbf{0.082} & \underline{0.781} & \textbf{0.921} & \textbf{0.083} & \underline{0.757} & \textbf{0.893} \\
\hline
\hline

 \multirow{4}{*}{BrNet \cite{han2022brnet}}  &   &    &   & 0.224 & 3.440 & 0.883 & 0.105 & 1.052 & 0.889 & 0.106 & 1.124 & 0.707 \\
& \checkmark &   &   &  0.168 & 2.041 & 0.914 & \underline{0.082} & 0.747 & 0.920 & 0.087 & 0.780 & 0.809 \\
& \checkmark & \checkmark &   &  \underline{0.119} & \underline{0.670} & \underline{0.918} & \textbf{0.080} & \underline{0.736} & \underline{0.922} & \underline{0.084} & \underline{0.722} & \underline{0.872} \\
& \checkmark & \checkmark & \checkmark &  \textbf{0.106} & \textbf{0.577} & \textbf{0.921} & \textbf{0.080} & \textbf{0.717} & \textbf{0.923} & \textbf{0.081} & \textbf{0.696} & \textbf{0.892} \\
\hline



\end{tabular}
\vspace{-1.0mm}
\caption{Results of training different depth estimation networks backbone under different settings. For each backbone, the best results are in \textbf{bold}, the second best are \underline{underlined}. \noindent\textbf{With pixel-wise:} The model is trained in our first stage only. 
\noindent\textbf{With pixel motion and pseudo depth:} Our fully self-supervised pipeline is applied to train the model.   
\noindent\textbf{With semantic prior:} Objects and ground masks are produced by a pre-trained segmentation model \cite{zou2023segment} instead of the self-supervised ones.}
\label{Tab:result_stage_wise_training}
\end{table*}

\begin{table*}[h]
\footnotesize
\centering
\begin{tabular}{|c|c|c|ccc|ccc|ccc|}
\hline
\multirow{2}{*}{Method} &
\multirow{2}{*}{Dataset} &
  \multirow{2}{*}{\begin{tabular}[c]{@{}c@{}}Sematic\\ prior\end{tabular}} &
  \multicolumn{3}{c|}{Dynamic region} &
  \multicolumn{3}{c|}{Static region} &
  \multicolumn{3}{c|}{All region} \\ \cline{4-12} 
   &
   &
   & 
  \multicolumn{1}{c|}{Abs Rel} &
  \multicolumn{1}{c|}{Sq Rel} &
  $\delta < 1.25$ &
  \multicolumn{1}{c|}{Abs Rel} &
  \multicolumn{1}{c|}{Sq Rel} &
  $\delta < 1.25$ &
  \multicolumn{1}{c|}{Abs Rel} &
  \multicolumn{1}{c|}{Sq Rel} &
  $\delta < 1.25$ \\ \hline \hline
Monodepth2* \cite{godard2019digging} &
\multirow{5}{*}{C}
        &
        &
  0.286 &
  6.036 &
  0.716 &
  0.119 &
  1.461 &
  0.877 &
  0.127 &
  1.678 &
  0.872 \\
Li \textit{et al.}* \cite{li2021unsupervised} & 
   &
   &
  0.188 &
  1.654 &
  0.735 &
  0.118 &
  1.198 &
  0.835 &
  0.119 &
  1.186 &
  0.833 \\
InstaDM \cite{lee2021learning} & 
    &
  \checkmark &
  0.189 &
  2.538 &
  0.795 &
  0.102 &
  1.058 &
  0.895 &
  0.106 &
  1.104 &
  0.890 \\
RMDepth \cite{hui2022rm} & 
   &
   &
  \textendash &
  \textendash &
  \textendash &
  \textendash &
  \textendash &
  \textendash &
  0.100 &
  0.839 &
  0.895 \\

DaCNN \cite{han2023self} & 
   &
   &
  \textendash &
  \textendash &
  \textendash &
  \textendash &
  \textendash &
  \textendash &
  0.113 &
  1.380 &
  0.888 \\

\hline

ResNet18 \cite{gordon2019depth}   + Ours & 
\multirow{4}{*}{C}
   &
   &
  \textbf{\underline{0.107}} &
  \underline{0.784} &
  \textbf{\underline{0.907}} &
  \underline{0.093} &
  \underline{0.961} &
  \underline{0.907} &
  \underline{0.094} &
  0.931 &
  \underline{0.906} \\
PackNet \cite{guizilini20203d} + Ours & 
   &
   &
  \underline{0.112} &
  \underline{0.703} &
  \underline{0.890} &
  \underline{0.090} &
  \underline{0.858} &
  \underline{0.909} &
  \underline{0.091} &
  \underline{0.831} &
  \underline{0.908} \\
DiffNet \cite{zhou2021self} + Ours & 
   &
   &
  \underline{0.113} &
  \textbf{\underline{0.657}} &
  \underline{0.889} &
  \underline{0.083} &
  \underline{0.775} &
  \underline{0.919} &
  \underline{0.085} &
  \underline{0.753} &
  \underline{0.916} \\
BrNet \cite{han2022brnet} + Ours & 
   &
   &
  \underline{0.119} &
  \underline{0.670} &
  \underline{0.872} &
  \textbf{\underline{0.080}} &
  \textbf{\underline{0.736}} &
  \textbf{\underline{0.922}} &
  \textbf{\underline{0.084}} &
  \textbf{\underline{0.722}} &
  \textbf{\underline{0.918}} \\
  \hline \hline 

PackNet \cite{guizilini20203d} & 
\multirow{5}{*}{K}
   &
   &
   0.213 &
   2.820 &
   0.762 &
   0.108 &
   0.814 &
   0.883 &
   0.110 &
   0.834 &
   0.881 \\  

DiffNet \cite{zhou2021self} & 
   &
   &
   0.177 &
   2.072 &
   0.930 &
   0.101 &
   0.743 &
   \textbf{0.899} &
   0.102 &
   0.753 &
   0.897 \\  

BrNet* \cite{han2022brnet} & 
   &
   &
   0.183 &
   1.715 &
   0.760 &
   0.104 &
   0.702 &
   0.885 &
   0.106 &
   0.711 &
   0.884 \\  
RMDepth \cite{hui2022rm} & 
   &
   &
  \textendash &
  \textendash &
  \textendash &
  \textendash &
  \textendash &
  \textendash &
   {0.108} &
   {0.710} &
   {0.884} \\ 
DaCNN \cite{han2023self} & 
   &
   &
  \textendash &
  \textendash &
  \textendash &
  \textendash &
  \textendash &
  \textendash &
   \textbf{0.099} &
   \textbf{0.661} &
   \textbf{0.897} \\   
\hline
DiffNet \cite{zhou2021self} + Ours & 
\multirow{4}{*}{K}
   &
   &
   \underline{0.158}&
   \underline{1.468} &
   \underline{0.838} &
   0.101 &
   \underline{0.687} &
   0.894 &
   0.102 &
   0.693 &
   0.892 \\  
DiffNet \cite{zhou2021self} + Ours & 
    & \checkmark
   &
   \textbf{\underline{0.143}} &
   \textbf{\underline{1.191}} &
   \textbf{\underline{0.845}} & 
   \textbf{\underline{0.099}} &
   \textbf{\underline{0.658}} &
   0.895 &
   0.100 &
   0.662 &
   0.894 \\ 
BrNet \cite{han2022brnet} + Ours & 
    &
   &
   \underline{0.160} &
   \underline{1.273} &
   \underline{0.812} &
   0.103 &
   \underline{0.690} &
   \textbf{0.889} &
   0.103 &
   0.692 &
   0.888 \\  

BrNet \cite{han2022brnet} + Ours & 
    & \checkmark
   &
   \underline{0.162} &
   \underline{1.207} &
   \underline{0.812} &
   0.102 &
   \underline{0.673} &
   \textbf{0.889} &
   0.103 &
   0.675 &
   0.888 \\  
\hline
\end{tabular}
\vspace{-1.0mm}
\caption{Comparison between the models trained with our method and previous works on Cityscapes (C) and Kitti (K) dataset. The best results are in \textbf{bold}. Additionally, the performance of our methods is \underline{underlined} if it is better than all previous works. (*) indicates results reproduced using the official code. All methods are trained on monocular videos.}
\label{Tab:result_comparision_to_SOTA}
\end{table*}

\label{Sec:4.3_stage_wise_training_subsection}
\vspace{-0.5mm}
\noindent\textbf{Performance Improvements:}
Our method improves the performance of various self-/unsupervised monocular depth estimation networks, especially in dynamic regions, on the Cityscapes dataset.
Specifically, we apply our method to a range of self-supervised monocular depth estimation network backbones, starting from the simple \emph{ResNet18} (with randomized layer normalization \cite{gordon2019depth}) to the more recent ones including \emph{PackNet} \cite{guizilini20203d}, \emph{DiffNet} \cite{zhou2021self} and \emph{BrNet} \cite{han2022brnet}. 
The performances of each network architecture under different settings are reported in Tab. \ref{Tab:result_stage_wise_training}.

The quantitative results indicate that when the models are trained using photometric loss only, errors in dynamic regions are considerably large (the first rows of each backbone). 
Integrating pixel-wise object motion prediction improves the performance; however, the depth predictions for dynamic regions are not scale-consistent due to the scale ambiguity (the second rows of each backbone). Our pseudo depth label provides scale-consistent depth supervision, which further improves depth predictions across all backbones (the third row of each  backbone).

The introduction of our models leads to an average of \textbf{52.1\%} reduction in the square relative error for the dynamic region, and \textbf{2.8\%} and \textbf{5.9\%} reduction for static and all regions respectively. Note the performance of the network on the static regions largely reflects the performance of the network on all regions as the static regions occupy the majority area of the input image. Therefore, the error reduction on  all-region is not as significant as that for dynamic regions only. We further show that by applying our method in conjunction with a pre-trained segmentation model, we reduce the square relative errors by \textbf{55.8\%}, \textbf{4.0\%}, \textbf{7.6\%} on average in dynamic, static, and all regions, respectively (the last rows of each backbone).







\begin{figure*}
    \footnotesize
    \centering
    \qquad  
    \setlength\tabcolsep{1.0pt}
    \begin{tabular}{c c c c c}
        \multirow{1}{*}[+6.5ex]{\rotatebox{90}{Image}} & \includegraphics[width=.24\textwidth]{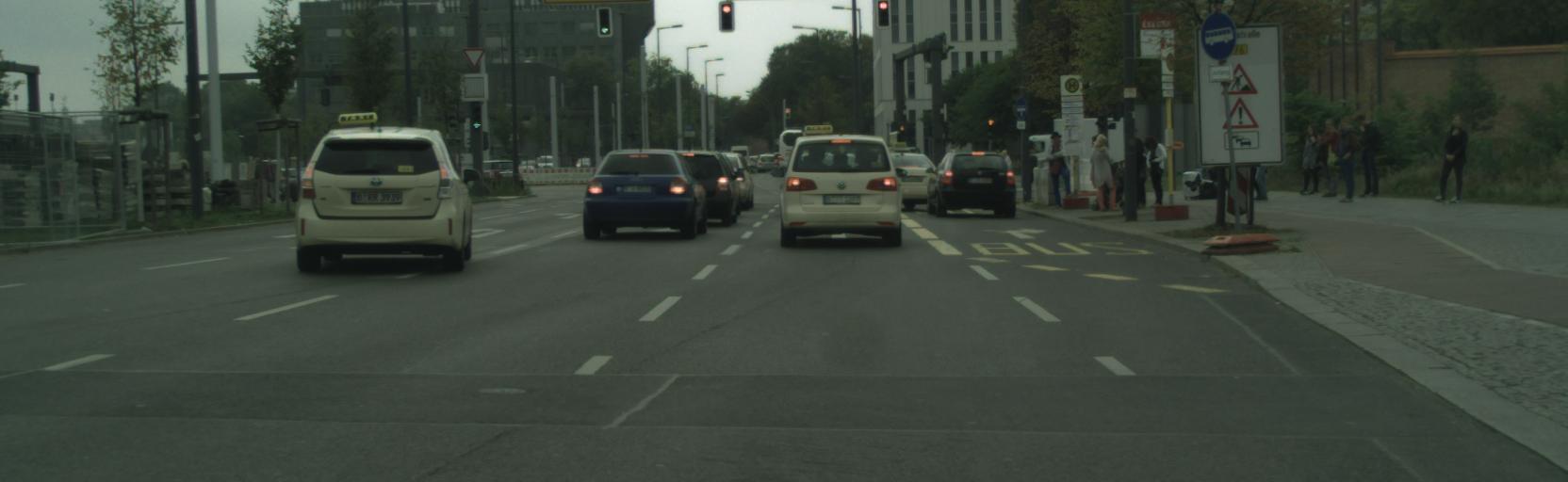} & \includegraphics[width=.24\textwidth]{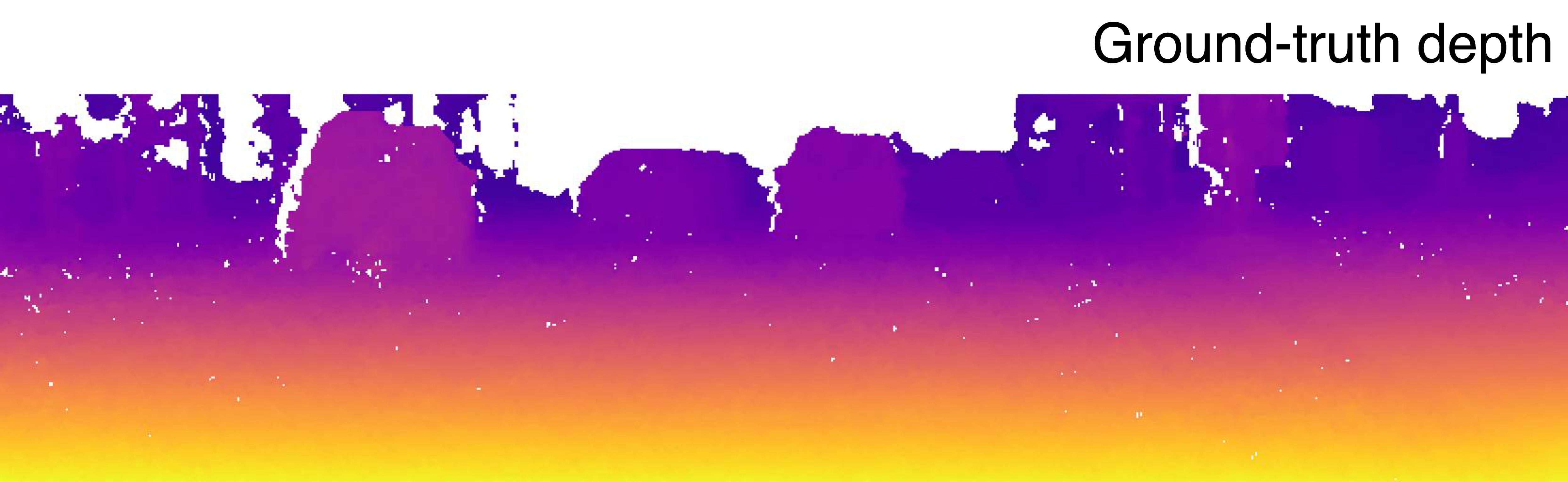} 
        & \includegraphics[width=.24\textwidth]{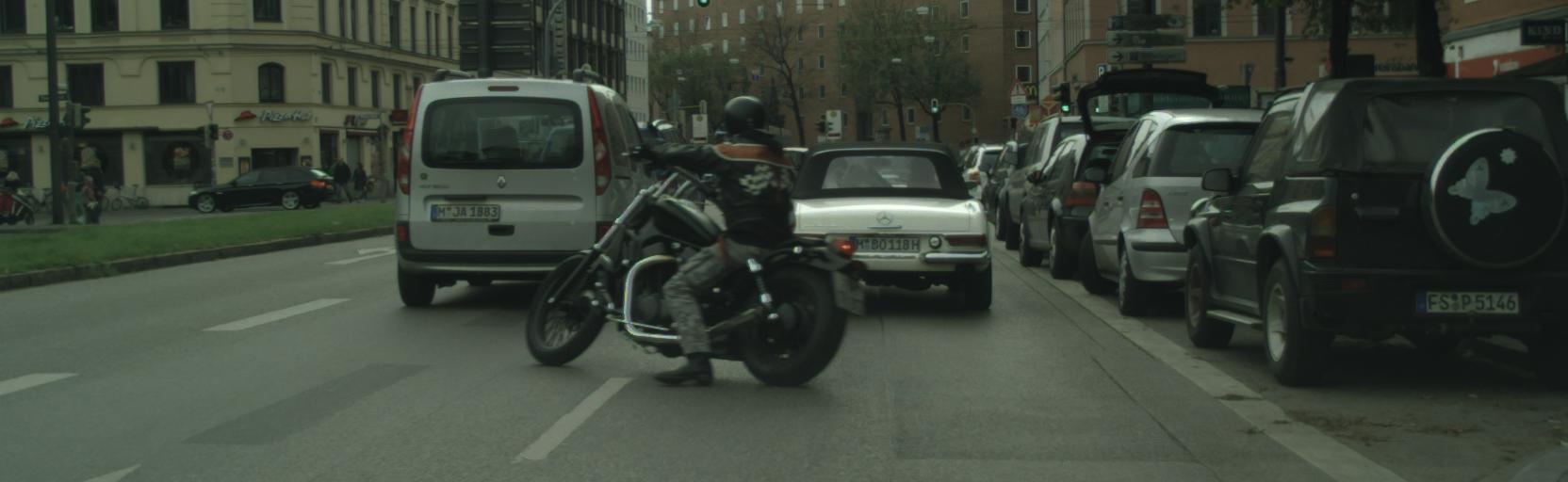} & \includegraphics[width=.24\textwidth]{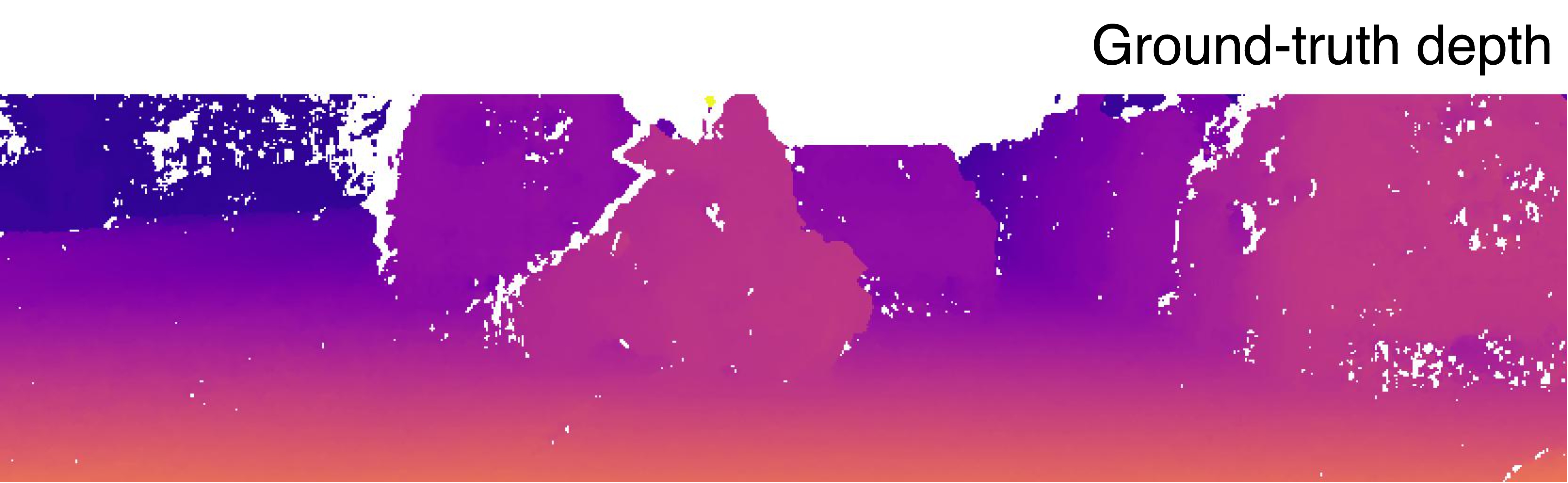} \\ 

        \multirow{1}{*}[+4.5ex]{\rotatebox{90}{\cite{li2021unsupervised}}} & \includegraphics[width=.24\textwidth]{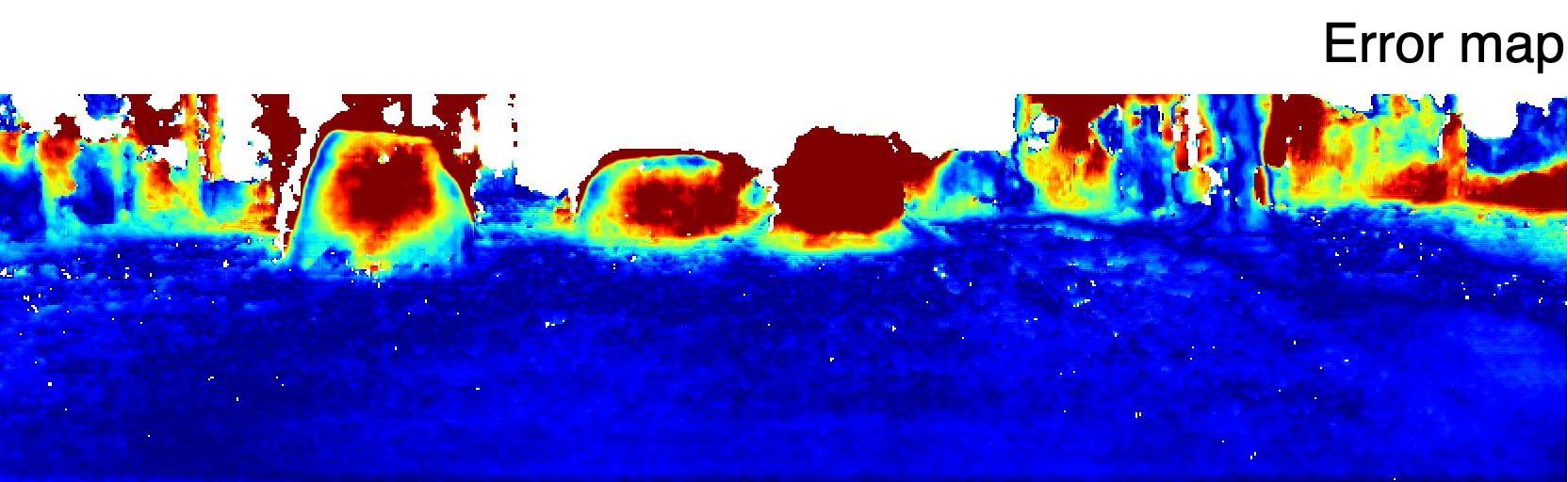} & \includegraphics[width=.24\textwidth]{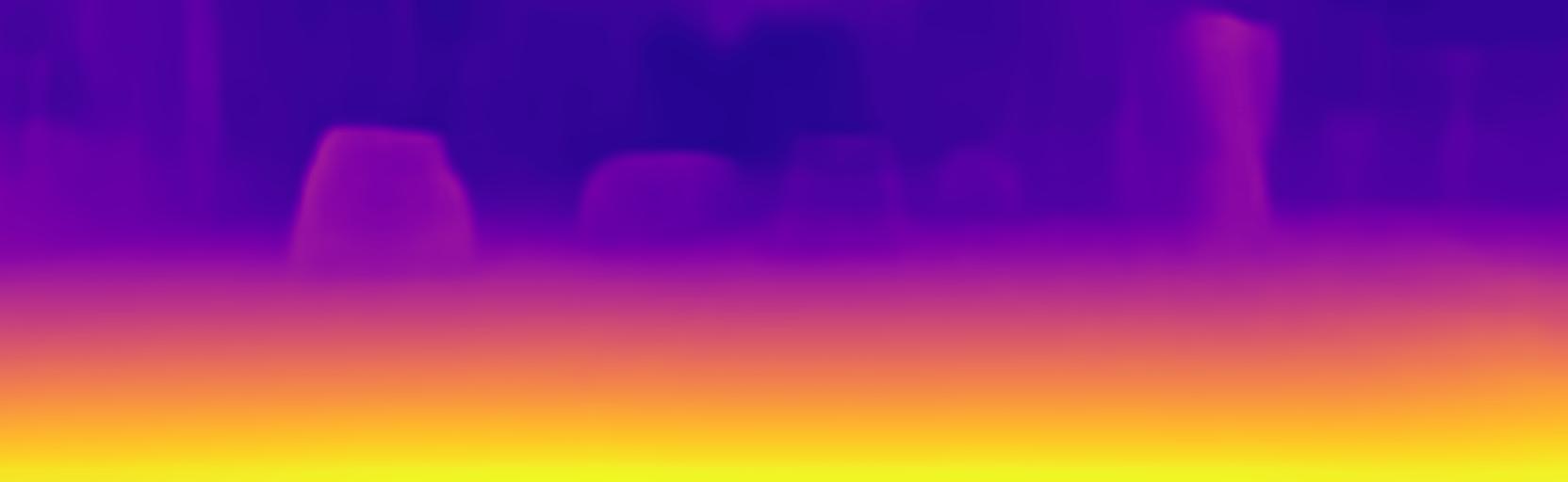} 
        & \includegraphics[width=.24\textwidth]{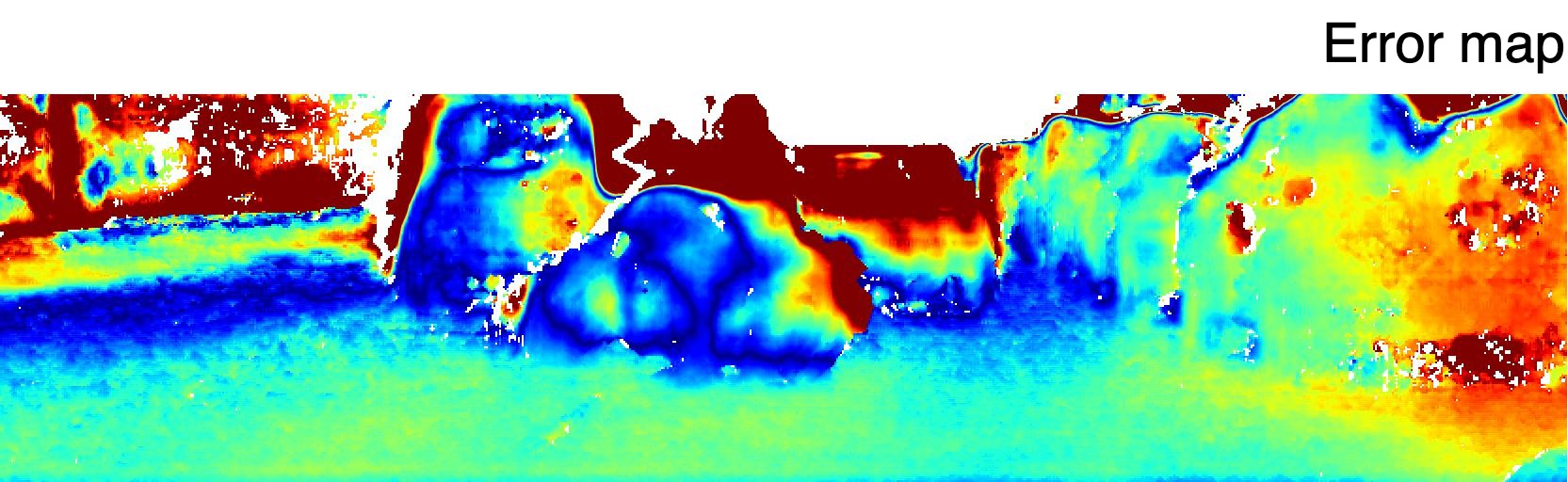} & \includegraphics[width=.24\textwidth]{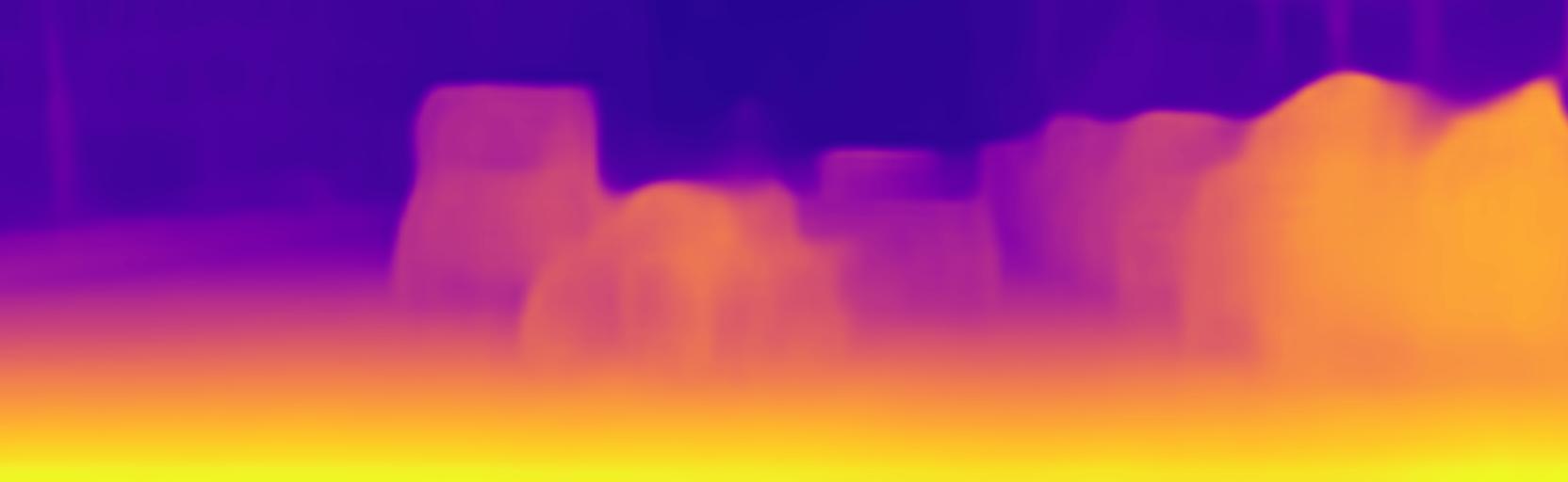} \\ 

        \multirow{1}{*}[+5.0ex]{\rotatebox{90}{\cite{lee2021learning}}} & \includegraphics[width=.24\textwidth]{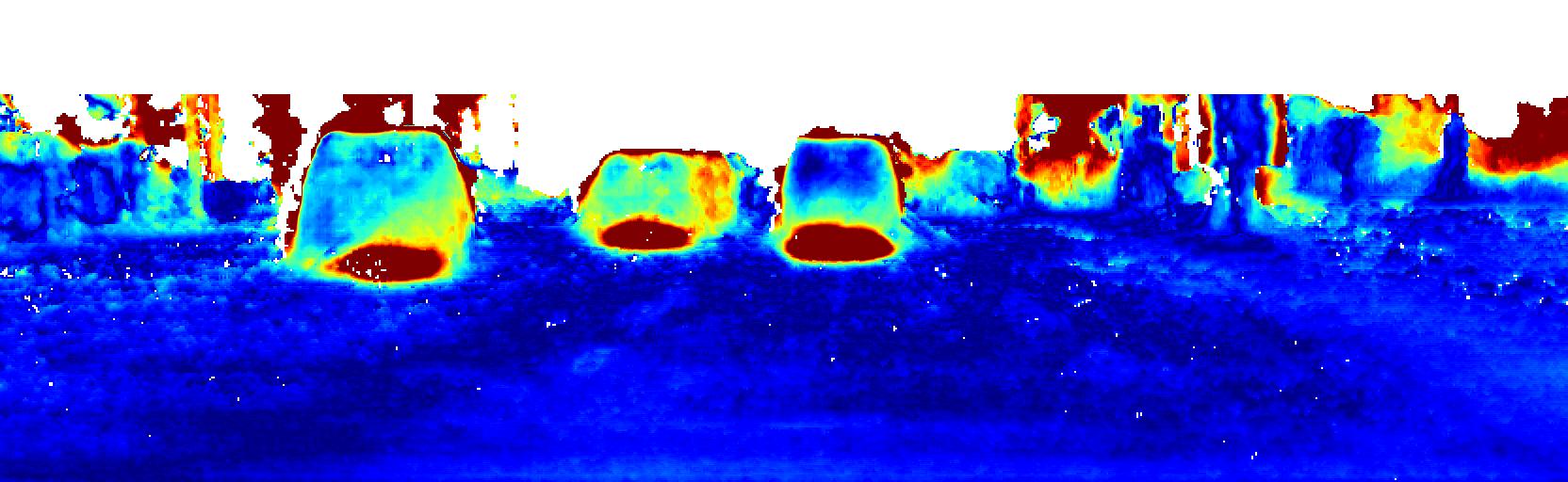} & \includegraphics[width=.24\textwidth]{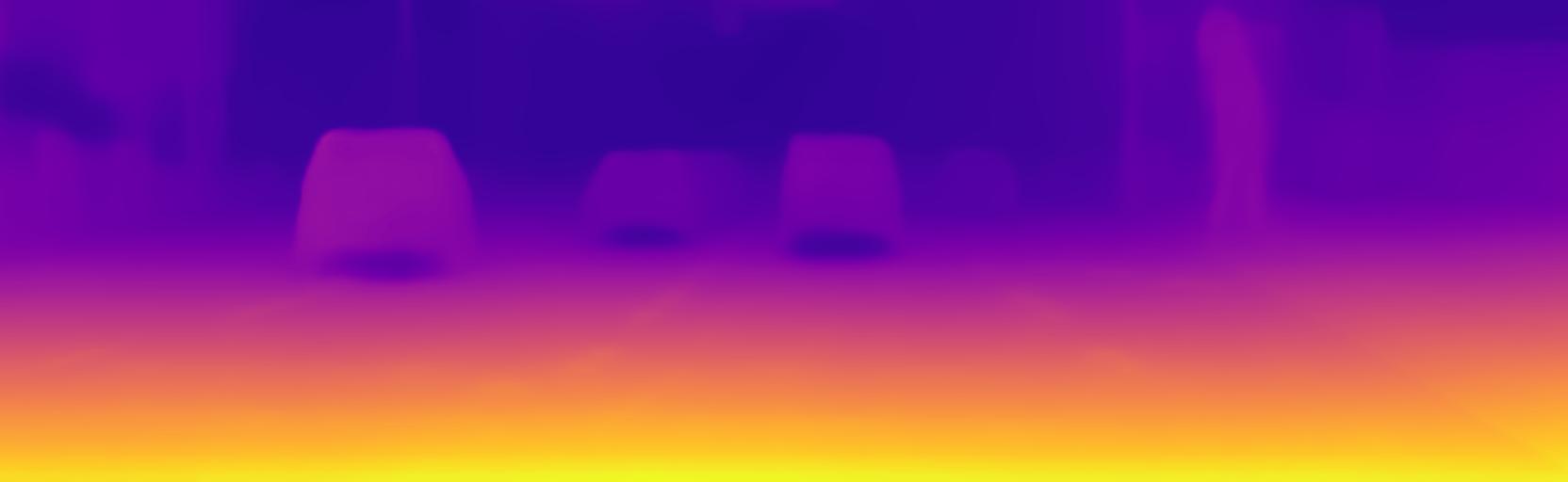} 
        & \includegraphics[width=.24\textwidth]{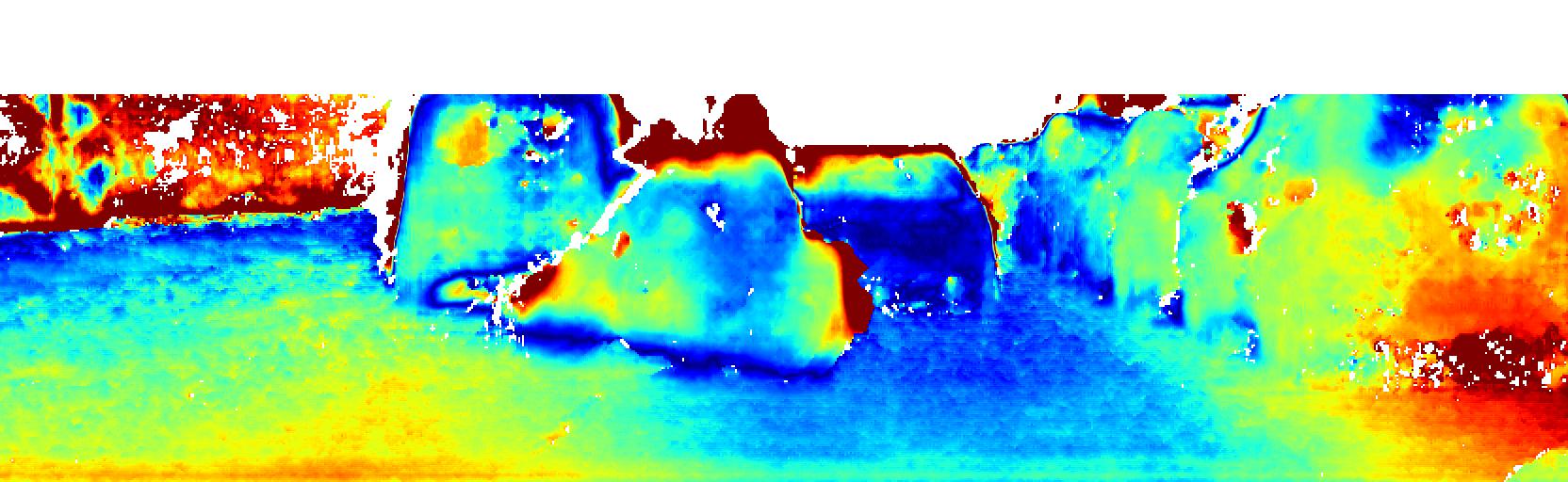} & \includegraphics[width=.24\textwidth]{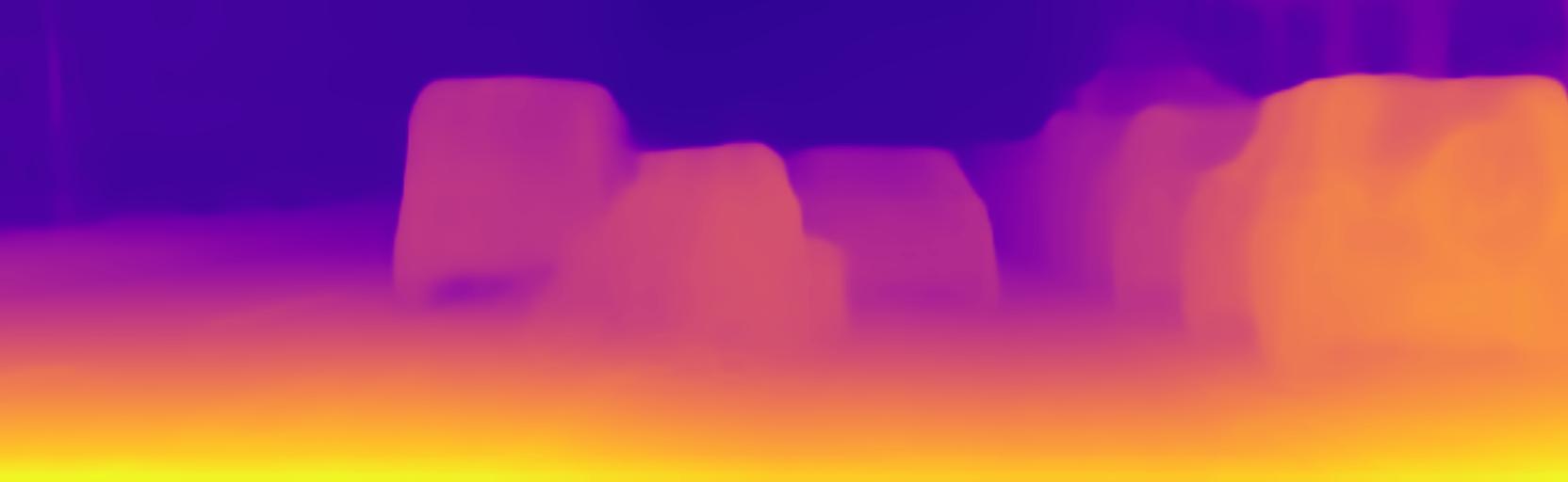} \\

        \multirow{1}{*}[+5.0ex]{\rotatebox{90}{Ours}} & \includegraphics[width=.24\textwidth]{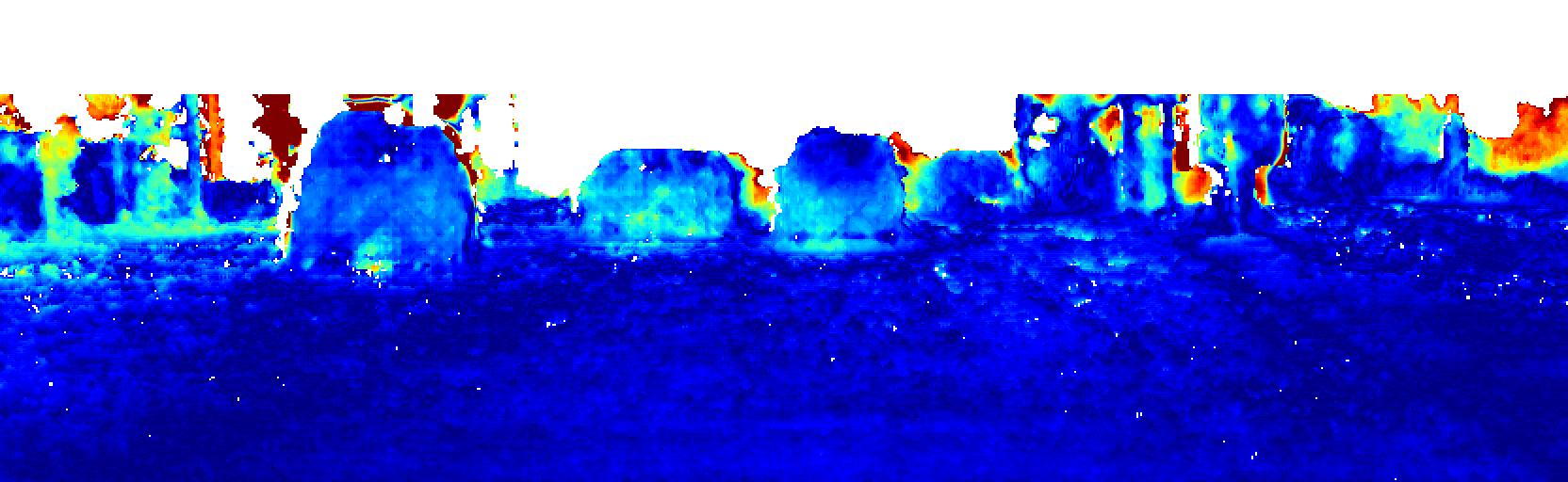} & \includegraphics[width=.24\textwidth]{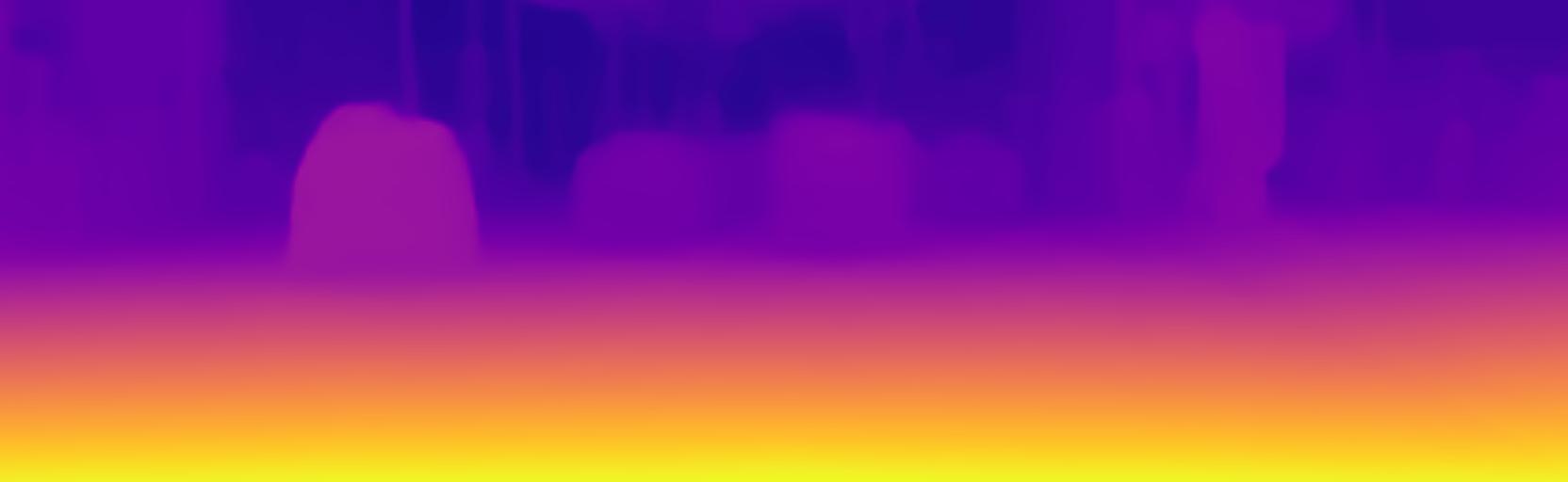} 
        & \includegraphics[width=.24\textwidth]{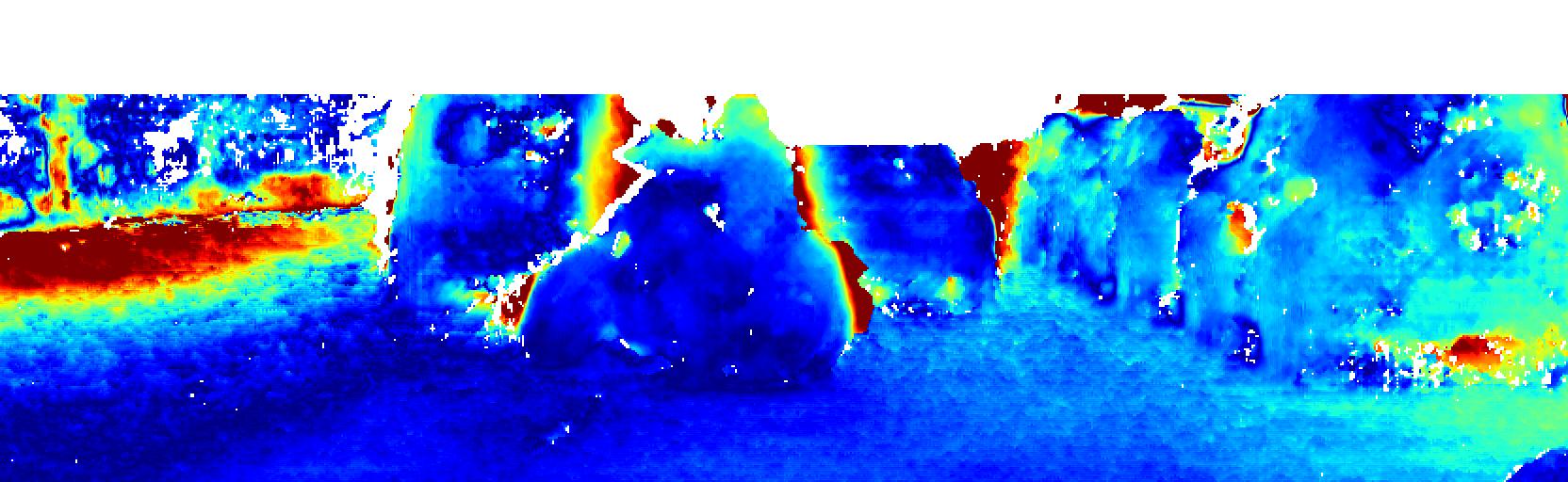} & \includegraphics[width=.24\textwidth]{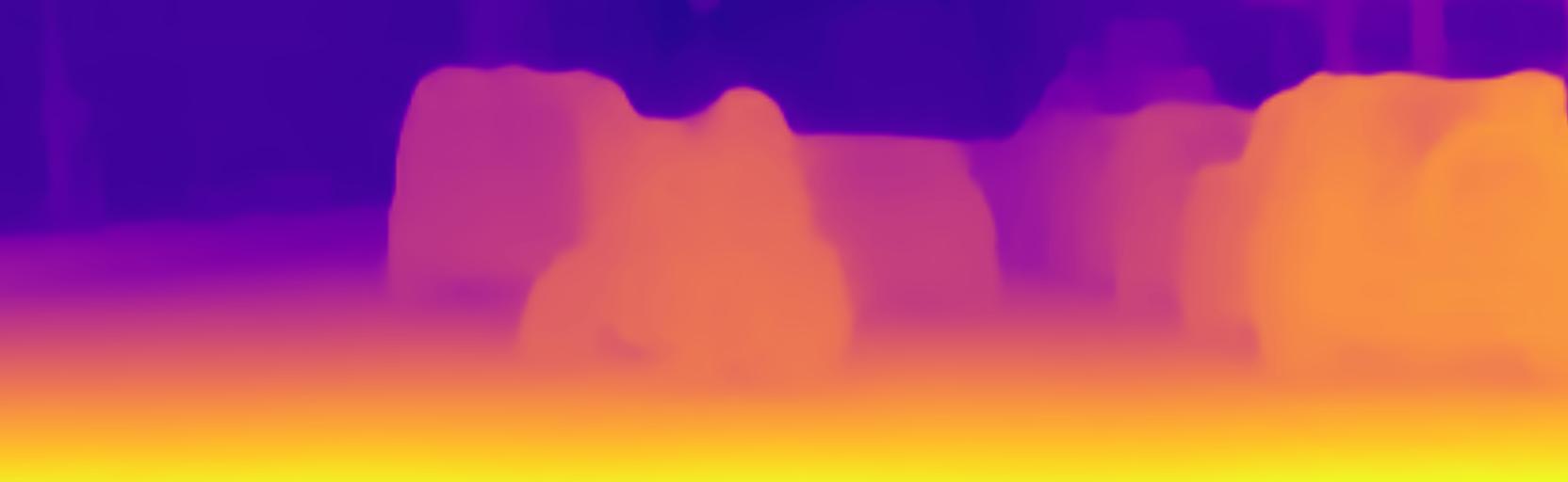} \\

        \multirow{1}{*}[8.0ex]{\rotatebox{90}{Ours + \cite{zou2023segment}}} & \includegraphics[width=.24\textwidth]{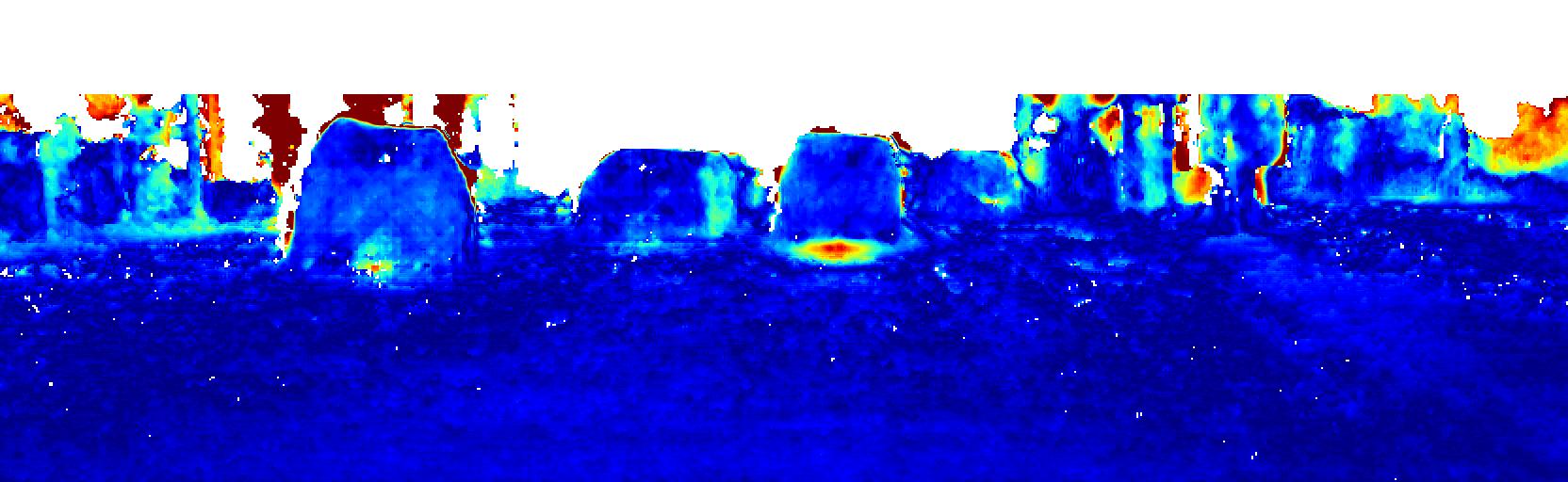} & \includegraphics[width=.24\textwidth]{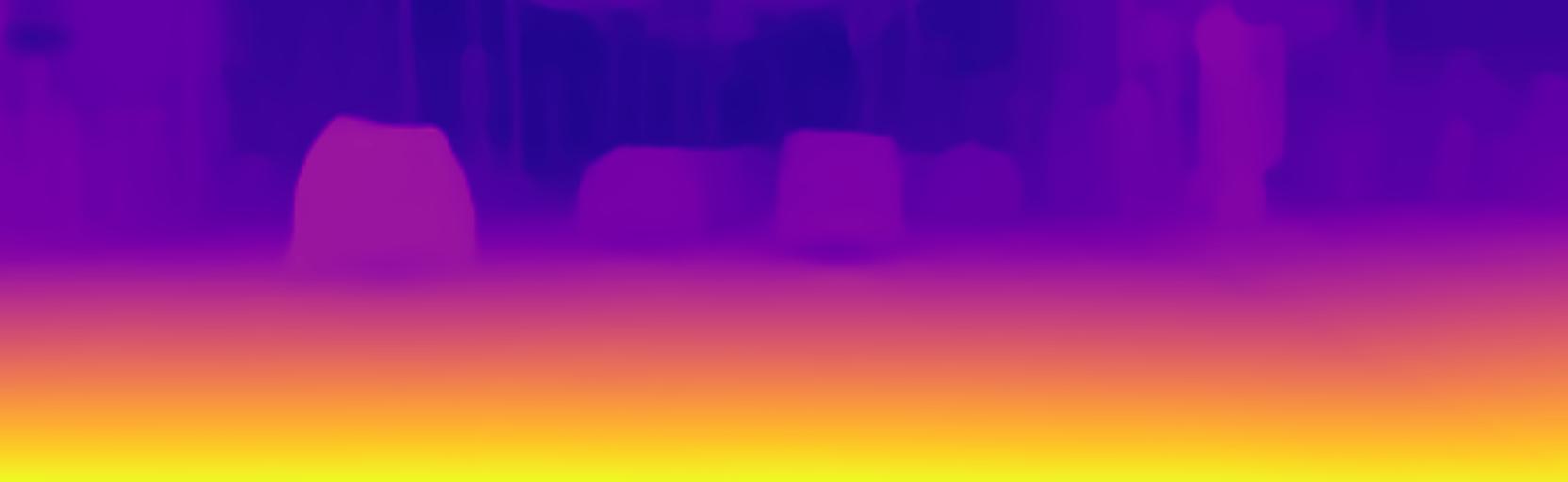} 
        & \includegraphics[width=.24\textwidth]{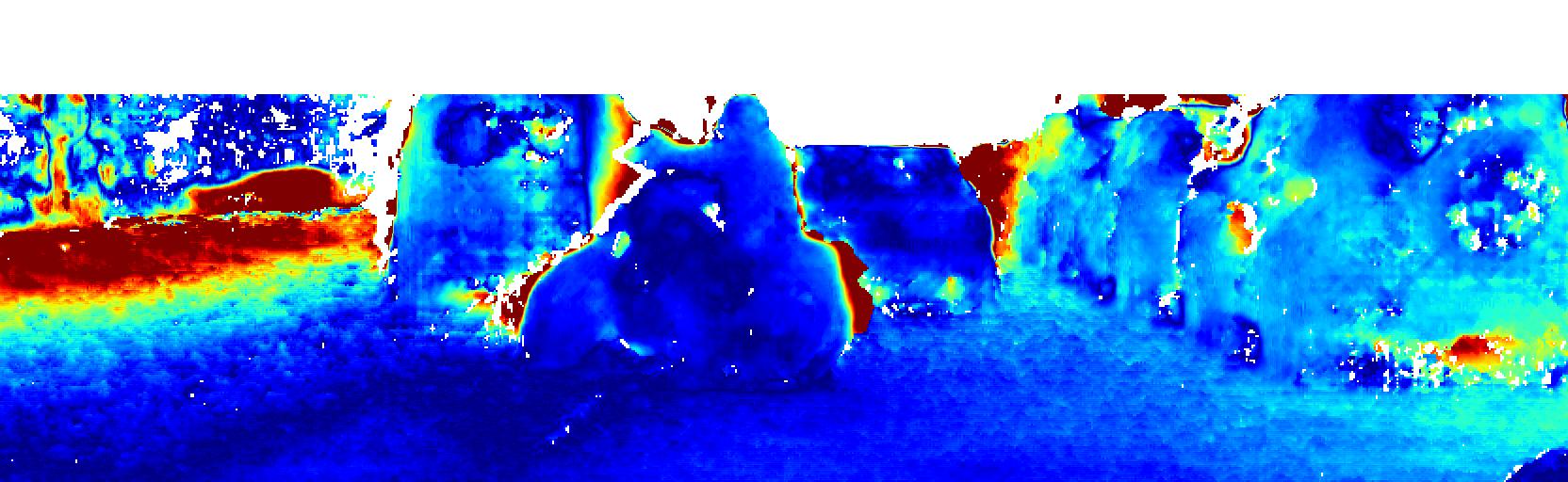} & \includegraphics[width=.24\textwidth]{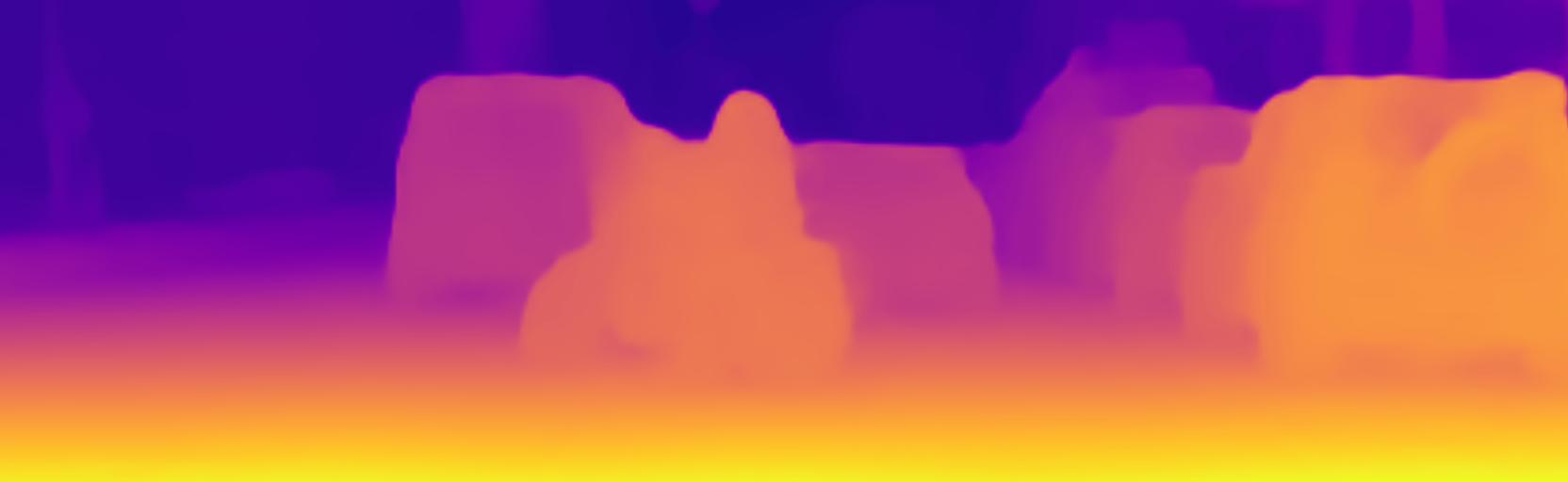}
   
    \end{tabular}
    \vspace{-1.0mm}
    \caption{Qualitative results on Cityscapes. The error map is computed as the per-pixel relative absolute depth error. Red areas indicate high errors. While the depth maps are not clearly distinguishable, the error maps indicate that both \cite{lee2021learning} and \cite{li2021unsupervised} fail to predict accurate depth for moving objects in the images. In contrast, our models are able to predict the depth for these objects with smaller errors. Our model, in conjunction with a pre-trained segmentation model \emph{SEEM} \cite{zou2023segment}, can achieve even better predictions.}
    \label{figure:4_qualitative_result}
\end{figure*}

\begin{figure*}
    \centering
    \begin{subfigure}{0.32\textwidth}
        \includegraphics[width=\linewidth]{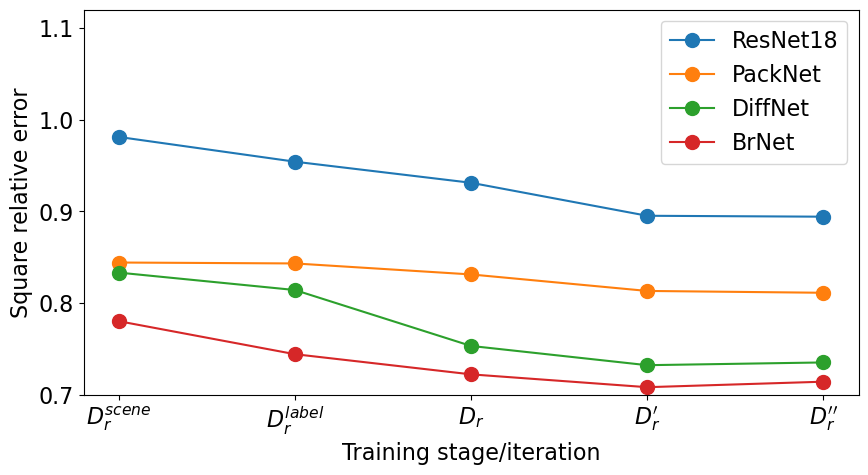}
        \caption{All region}
    \end{subfigure}
    \hfill
    \begin{subfigure}{0.32\textwidth}
        \includegraphics[width=\linewidth]{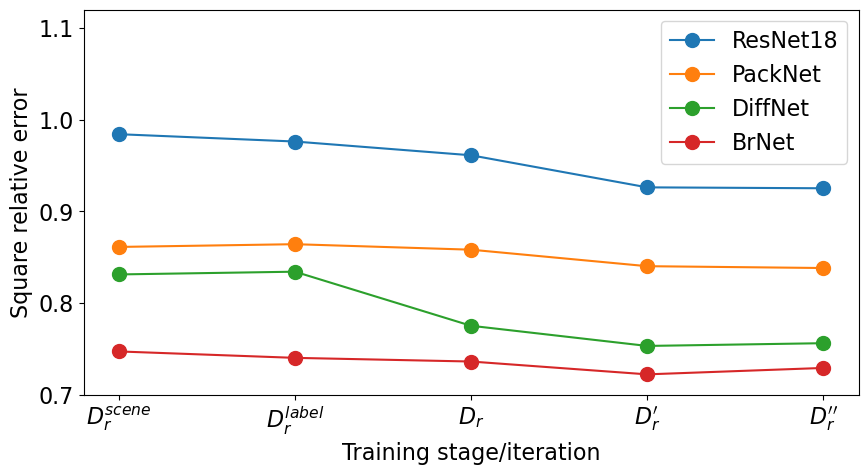}
        \caption{Static region}
    \end{subfigure}
    \hfill
    \begin{subfigure}{0.32\textwidth}
        \includegraphics[width=\linewidth]{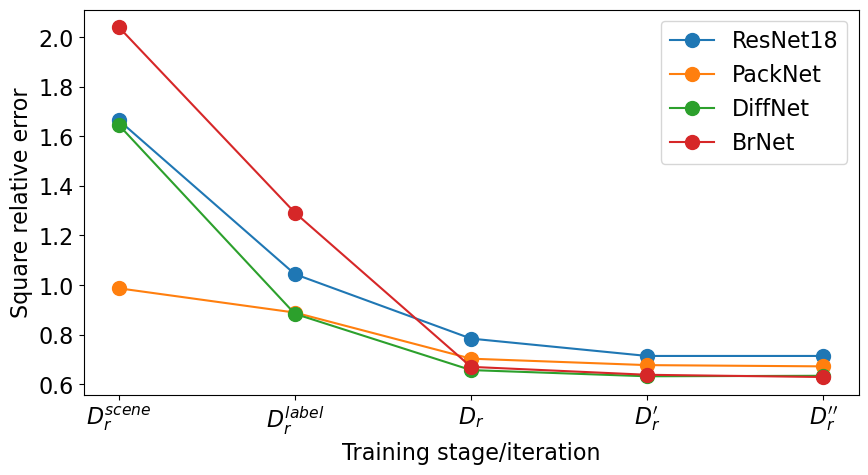}
        \caption{Dynamic region}
    \end{subfigure}
    \vspace{-2.0mm}
    \caption{Depth prediction errors at different stage/self-training iteration. The largest performance gains are achieved while fine-tuning our method with the pseudo label $\mathbf{D}_r^{label}$. Their performance slightly improves through our iterative training strategy.}
        
    \label{fig:4.5_iterative_training_curve}
\end{figure*}

\noindent\textbf{Baseline Comparison:}
In Table \ref{Tab:result_comparision_to_SOTA}, we compare our proposed framework against prior methods on the Cityscapes and KITTI dataset quantitatively. For works with no official pre-trained model release \cite{godard2019digging, li2021unsupervised, han2022brnet}, we re-train the models with their official implementation and evaluate on static and dynamic regions separately.
For \cite{hui2022rm, {han2023self}} we only report the all-region performance as there is no publicly available implementation or pre-trained models. 
For the rest works, we evaluate their pre-trained models directly. 
All models are evaluated following the process detailed in \cite{watson2021temporal}.

The results show that, in the Cityscapes dataset, our models outperform all the previous methods on dynamic regions by significant margins across all metrics. For the Sq Rel metric, our models outperform the best baseline by a margin of $\mathbf{0.870}$, that is, a \textbf{52.6\%} error reduction. By resolving scale ambiguity under the presence of moving objects, our models achieve new state-of-the-art performance on the Cityscapes dataset. To be more specific, the model with \emph{ResNet18} backbone achieves state-of-the-art on the Abs Rel and Sq Rel metrics. Our other models outperform all previous works in all metrics across different types of regions. The improvement of the performance are even larger when a pre-trained segmentation model is available. 
Figure \ref{figure:4_qualitative_result} shows the qualitative results on the Cityscapes dataset. Although \cite{lee2021learning}
 and \cite{li2021unsupervised} produce low-error predictions for static regions, they fail to predict accurate depth for the moving objects. In contrast, our methods are able to accurately estimate depth for both static and dynamic regions. 

For KITTI dataset, our models outperform all previous works in dynamic regions and achieve a reduction of at least \textbf{14.4\%} in the Sq Rel metric. Due to a small number of moving objects in the image, scale-consistent depth predictions in dynamic regions result in marginal all-region improvements. While our models do not outperform \emph{DaCNN}\cite{{han2023self}} in all-region\footnote{We cannot combine our training method with \emph{DaCNN} \cite{han2023self} as  is not publicly available.}, we expect to achieve higher performance by combining our training method with their network design.

\vspace{-1.5mm}
\subsection{Iterative Self-training}
\vspace{-1.0mm}

In this section, we investigate, as a natural extension to our pipeline, a self-training setup where we use the depth prediction from $\Theta$ as the pseudo depth label and finetune $\Theta$ itself following the procedure described in Sec. \ref{section:final_depth_prediction}.
In Fig. \ref{fig:4.5_iterative_training_curve}, we visualize the change of errors of ours models at different stage/self-training iteration. In the figure, we use $\mathbf{D}'_r$ and $\mathbf{D}^{''}_r$  to represent the depth prediction after one and two self-training iterations. 
The figure shows that the largest performance gain happens at fine-tuning with pseudo label $\mathbf{D}_r^{label}$.
The performance of our models slightly improve at the first iteration and remains stable after that.

\vspace{-1.5mm}
\section{Conclusion}
\vspace{-1.0mm}
In this work, we proposed a fully self-supervised pipeline that generates scale-consistent depth pseudo labels. Our decoupled depth estimation for  static and dynamics regions can lead to a high-quality pseudo depth labels. Such labels have been used to improve the performance of a range of self-supervised monocular depth estimation frameworks, especially in regions with moving objects. Extensive experimental results demonstrate that  we achieve the new state-of-the-art performance on the Cityscapes dataset, and consistently outperform previous works in depth predictions for dynamic regions on both KITTI and Cityscapes dataset. One limitation of our method is its generalizability to indoor scenes in which the main moving objects are humans whose motion cannot be described by a single rigid motion.

\vspace{-4.0mm}
\noindent\paragraph{Acknowledgement.}
This research was supported in part by the Australia Research Council DECRA Fellowship (DE180100628) and ARC Discovery Grant (DP200102274)).







{
    \small
    \bibliographystyle{ieeenat_fullname}
    \bibliography{main}
}


\end{document}